\documentclass{article}

\PassOptionsToPackage{numbers, compress}{natbib}
\usepackage[final]{neurips_2022}

\usepackage[utf8]{inputenc} 
\usepackage[T1]{fontenc}    
\usepackage{hyperref}       
\usepackage{url}            
\usepackage{booktabs}       
\usepackage{amsfonts}       
\usepackage{nicefrac}       
\usepackage{microtype}      
\usepackage{xcolor}         
\usepackage{multirow}
\usepackage{etoolbox}
\newcommand{\zerodisplayskips}{%
  \setlength{\abovedisplayskip}{0pt}%
  \setlength{\belowdisplayskip}{0pt}%
  \setlength{\abovedisplayshortskip}{0pt}%
  \setlength{\belowdisplayshortskip}{0pt}}
\appto{\normalsize}{\zerodisplayskips}
\appto{\small}{\zerodisplayskips}
\appto{\footnotesize}{\zerodisplayskips}

\usepackage{amsmath,graphicx}
\usepackage[colorinlistoftodos,bordercolor=orange,backgroundcolor=orange!20,linecolor=orange,textsize=scriptsize]{todonotes}

\title{GAPX: Generalized Autoregressive Paraphrase-Identification X}

\author{Yifei Zhou\\
  Cornell University  \\
  \texttt{yz639@cornell.edu} \\\And
  Renyu Li \\
  Cornell University \\
  \texttt{rl626@cornell.edu} \\\And
  Hayden Housen \\
  Cornell University  \\
  \texttt{hth33@cornell.edu} \\\And
  Ser-nam Lim\\
  Meta AI\\
  \texttt{sernamlim@fb.com} \\}

\begin{document}
\maketitle
\begin{abstract}
Paraphrase Identification is a fundamental task in Natural Language Processing. While much progress has been made in the field, the performance of many state-of-the-art models often suffer from distribution shift during inference time. We verify that a major source of this performance drop comes from biases introduced by negative examples. To overcome these biases, we propose in this paper to train two separate models, one that only utilizes the positive pairs and the other the negative pairs. This enables us the option of deciding how much to utilize the negative model, for which we introduce a perplexity based out-of-distribution metric that we show can effectively and automatically determine how much weight it should be given during inference. We support our findings with strong empirical results. \footnote{Our code is publicly available at: https://github.com/YifeiZhou02/generalized\_paraphrase\_identification}
\end{abstract}

\section{Introduction}
Paraphrases are sentences or phrases that convey the same meaning using different wording, and is fundamental to the understanding of languages \citep{10.1162/COLI_a_00166}. Paraphrase Identification is a well-studied task of identifying if a given pair of sentences has the same meaning \citep{DBLP:journals/corr/WangMI16a, tomar2017neural, DBLP:journals/corr/abs-1908-11828, yin2015convolutional, 9414944}, and has many important downstream applications such as machine translation \citep{DBLP:journals/corr/abs-1904-09675, DBLP:journals/corr/abs-2004-04696, DBLP:journals/corr/abs-2009-09025, kozareva2006paraphrase}, and question-answering \citep{cer-etal-2017-semeval, explorations-in-sentence}.

\begin{figure}[h!]
    \centering
    \includegraphics[width=\textwidth]{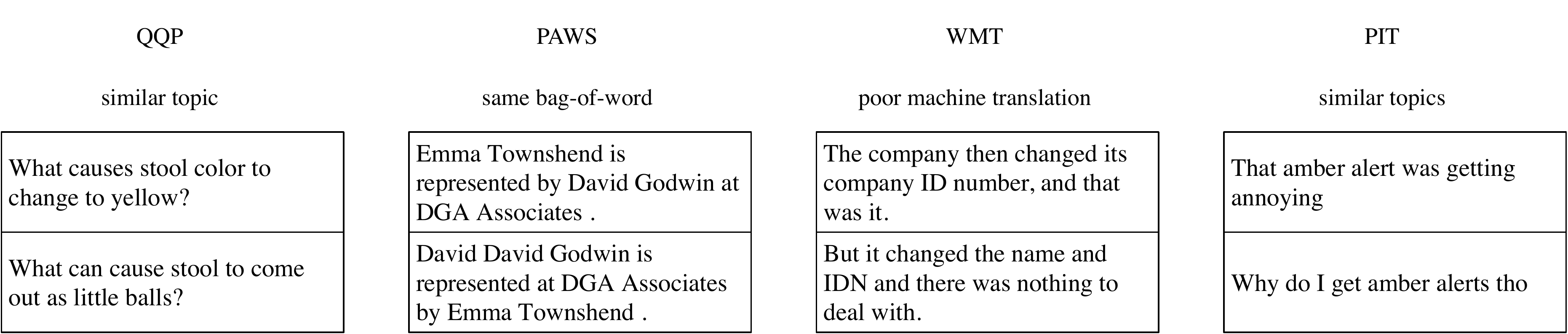}
    \caption{Negative pairs in different datasets are mined differently in different datasets, and can lead to significant biases during training.}
    \label{linregintro}
\end{figure}

Recently, researchers have observed that neural network architectures trained on different datasets could achieve state-of-the-art performances for the task of paraphrase identification \citep{DBLP:journals/corr/WangHF17, DBLP:journals/corr/abs-1810-04805, DBLP:journals/corr/abs-2104-14690}. While these advances are encouraging for the research community, it has however been observed that these models can be especially fragile in the face of distribution shift \citep{shen-lam-2021-towards}. In other words, when a model trained on a source dataset $\mathcal{D}^s$ is tested on another dataset, $\mathcal{D}^t$, collected and annotated independently, and with a distribution shift, the classification accuracy drops significantly \citep{DBLP:journals/corr/abs-1904-01130}. 

This paper presents our findings and observations that negative pairs (i.e., non-paraphrase pairs) in the training set, as opposed to the positive pairs, do not generalize well to out-of-distribution test pairs. Intuitively, negative pairs only represent a limited perspective of how the meanings of sentences can be different (and indeed it is practically infeasible to represent every possible perspective). We conjecture that negative pairs are so specific to the dataset that they adversely encourage the model to learn biased representations. We show this observation in Figure~\ref{linregintro}. Quora Question Pair (QQP) \footnote{https://quoradata.quora.com/First-Quora-Dataset-Release-Question-Pairs} extracts its negative pairs from similar topics. Paraphrase Adversarials from Word Scrambling (PAWS) \citet{DBLP:journals/corr/abs-1904-01130} generate negative pairs primarily from word swapping. World Machine Translation Metrics Task 2017 (WMT) \citep{Bojar2017ResultsOT} considers negative examples as poor machine translations. We therefore hypothesize that biases introduced by the different ways negative pairs are mined are major causes of the poor generalizability of paraphrase identification models.

Based on this observation, we would like to be able to control the reliance on negative pairs for out-of-distribution prediction. In order to achieve this, we propose to explicitly train two separate models for the positive and negative pairs (we will refer to them as the positive and negative models respectively), which will give us the option to choose when to use the negative model. It is well known that just training on positive pairs alone can lead to a degenerate solution~\cite{DBLP:journals/corr/abs-2006-07733, DBLP:journals/corr/abs-2103-03230, DBLP:journals/corr/abs-1908-10084, DBLP:journals/corr/abs-2002-05709}, e.g., a constant function network would still produce a perfect training loss. To prevent this, we propose a novel generative framework where we use an autoregressive transformer \citep{radford2019language, DBLP:journals/corr/VaswaniSPUJGKP17}, specifically BART \citep{DBLP:journals/corr/abs-1910-13461}. Given two sentences, we condition the prediction of the next token in the second sentence on the first sentence and the previous tokens. In a Bayesian sense, this would mean that the next token predicted has a higher probability of being a positive/negative pair to the first sentence for the positive and negative model respectively. This learning strategy has no degenerate solutions even when we are training the positive and negative models separately. We call our proposed approach GAP to stand for Generalized Autoregressive Paraphrase-Identification. One potential pitfall of GAP is that it ignores the ``interplay'' between positive and negative pairs that would otherwise be learned if they are utilized in training together. This is especially important when the test pairs are in-distribution. To overcome, we utilize an extra discriminative model, trained with both positive and negative pairs, to capture the interplay. We call this extension GAPX (pronounced as ``Gaps'') to capture the e\emph{X}tra discriminative model used.

For all practical purposes, the weights we placed on the positive, negative and/or discriminative model in GAP and GAPX need to be determined automatically during inference. For in-distribution pairs, we desire to use them all, while for out-of-distribution pairs, we hope to rely on the positive model much more heavily. This obviously leads to a question of how to determine whether a given pair is in or out of distribution \citep{9144212, choi2019waic, NEURIPS2019_1e795968, https://doi.org/10.48550/arxiv.1802.04865, NEURIPS2021_3941c435}. During testing, our method ensembles the positive model, the negative model, and the discriminative model based on the degree of the similarity of the test pair to the training pairs, and found that this works well for our purpose. We measure this degree of similarity with probability cumulative density function (cdf) in terms of perplexity \citep{DBLP:journals/corr/abs-1812-04606}, and show that it is superior to other measures.

In summary, our contributions are as follow:
\begin{enumerate}
    \item We report new research insights, supported by empirical results, that the negative pairs of a dataset could potentially introduce biases that will prevent a paraphrase identification model from generalizing to out-of-distribution pairs.
    \item To overcome,  we propose a novel autoregressive modeling approach to train both a positive and a negative model, and ensemble them automatically during inference. Further, we observe that the interplay between positive and negative pairs are important for in-distribution inference, for which we add a discriminative model. We then introduce a new perplexity based approach to determine whether a given pair is out-of-distribution to achieve auto ensembling.
    \item We support our proposal with strong empirical results. Compared to state-of-the-art transformers in out-of-distribution performance, our model achieves an average of 11.5\% and 8.4\% improvement in terms of macro F1 and accuracy respectively over 7 different out-of-distribution scenarios. Our method is especially robust to paraphrase adversarials like PAWS, while keeping comparable performance for in-distribution prediction.
\end{enumerate}
\section{Related Works} 
\subsection{Distribution Shift and Debiasing Models in NLP}
The issue of dataset bias and model debiasing has been widely studied in a lot of field in NLP such as Natural Language Inference \citep{he2019unlearn, amirkhani2020farstail} and Question Answering \citep{min-etal-2019-compositional, agrawal2018don, DBLP:journals/corr/abs-1811-05013}. Notable work by \citep{DBLP:journals/corr/abs-1908-10763, pmlr-v119-bahng20a, DBLP:journals/corr/abs-1906-10169, DBLP:journals/corr/abs-1909-03683, DBLP:journals/corr/abs-2009-12303} utilize ensembling to reduce models' reliance on dataset bias. These models share the same paradigm where they break down a given sample $x$ into signal feature $x_s$ and biased feature $x_b$, in the hope of preventing their model from relying on $x_b$, which has been shown to be the limiting factor preventing the model from generalizing to out-of-distribution samples \cite{pmlr-v119-bahng20a}. Here, a separate model is first either trained on $x_b$ or on datasets with known biases \citep{DBLP:journals/corr/abs-1906-10169, DBLP:journals/corr/abs-1909-03683, DBLP:journals/corr/abs-1908-10763}, or acquired from models known to have limited generalization capability. Then they train their main model with a regularization term that encourages the main model to produce predictions that deviate from that of the ``biased model''. However, this type of approach has shown limited success \citep{shen-lam-2021-towards} in debiasing paraphrase identification models. In contrast, our method is based on our observation that negative pairs limit the generalization of paraphrase identification models.

\subsection{Out-of-distribution Detection}
Another line of work relevant to this paper is the task of detecting out-of-distribution samples \citep{9144212,  choi2019waic, NEURIPS2019_1e795968, https://doi.org/10.48550/arxiv.1802.04865, NEURIPS2021_3941c435, DBLP:journals/corr/abs-2110-11334}. Researchers have proposed methods to detect anomaly samples by examining the softmax scores \citep{DBLP:journals/corr/LiangLS17} or energy scores \citep{DBLP:journals/corr/abs-2010-03759, DBLP:journals/corr/ZhaiCLZ16} produced by discriminative models, while others take a more probabilistic approach to estimate the probability density \citep{choi2019waic, DBLP:journals/corr/abs-1812-04606, zong2018deep, abati2019latent} or reconstruction error \citep{NEURIPS2018_5421e013}. In this paper, we introduce a novel perplexity based out-of-distribution detection method that we show empirically to work well for our purpose. Specifically, during inference, an out-of-distribution score is utilized to weigh the contributions from the positive and negative models: the higher the confidence that the sample is out-of-distribution, the lesser the negative model's contribution.

\subsection{Text Generation Metrics}
Finally, we would like to note the difference between our work and autoregressive methods that have been explored \citep{DBLP:journals/corr/abs-2106-11520, DBLP:journals/corr/abs-2004-14564} for evaluating text generation. Our work differs as follows: 1) Paraphrase identification seeks to assign a label of paraphrase or not while text generation metrics seeks to assign a score to measure the similarity of sentences; 2) Current text generation metrics either cannot be trained to fit to a specifc distribution \citep{DBLP:journals/corr/abs-1904-09675, DBLP:journals/corr/abs-2106-11520, DBLP:journals/corr/abs-1909-02622} or are limited to the i.i.d. setting \citep{DBLP:journals/corr/abs-2004-04696, rei-etal-2020-comet} of the training distribution. In contrast, our method not only significantly improves out-of-distribution performances but is also competitive with state-of-the-art paraphrase identification methods for in-distribution predictions.

\section{Methodology}
We observe that negative pairs in paraphrase identification constitute the main source of bias. To overcome this, we propose the following training paradigms to learn a significantly less biased paraphrase identification model. We employ an autoregressive conditional sentence generators with transformer architecture as the backbone of our model. Specifically, we train a positive and negative model to estimate the distribution of positive and negative pairs in a dataset respectively. During testing, the two models are ensembled based on how likely the input pair is out of distribution. This section provides details on our method.

\begin{figure*}[h!]
    \centering
    \includegraphics[width=\textwidth]{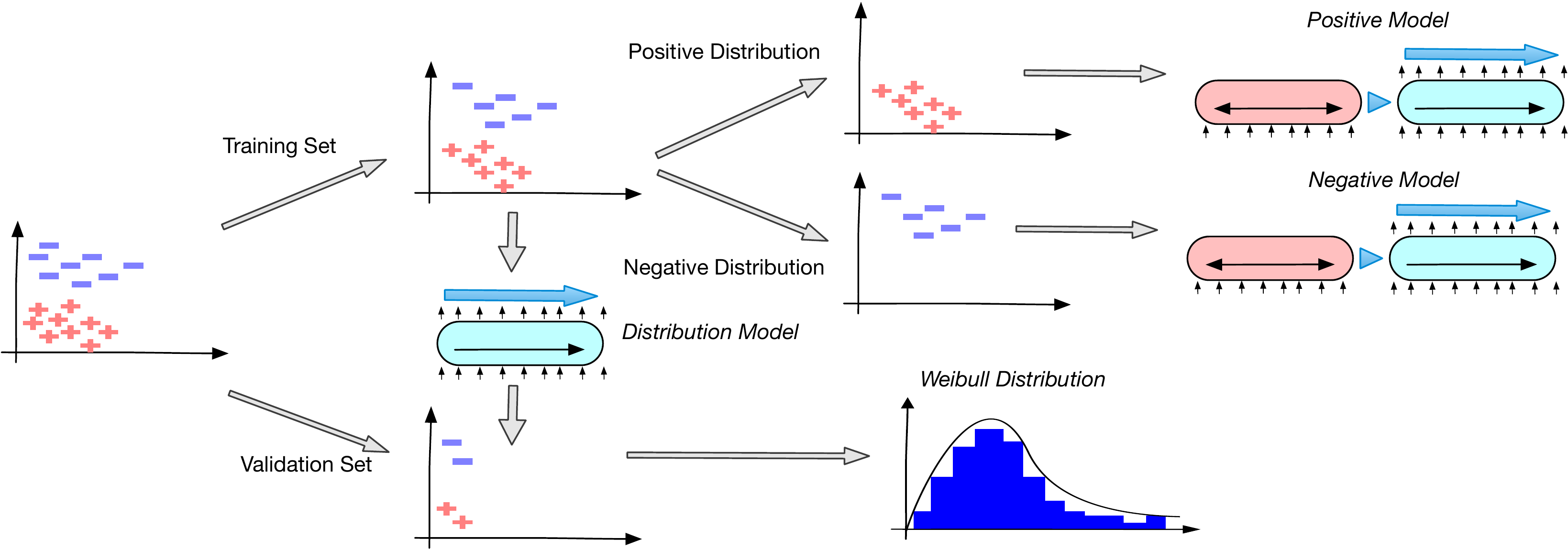}
    \caption{An overview of the training procedure of our model GAP. GAPX ensembles GAP with another discriminative model.}
    \label{overview}
\end{figure*}

\subsection{Separation of Dependence on Positive and Negative Pairs} \label{separation}
Let $\mathcal{S}$ be the space of all sentences, $X = (s_1, s_2)$ be the random variable representing a sample pair from $\mathcal{S}$, and $Y$ the random variable representing the labels, with $Y = 1$ indicating that $s_1$ and $s_2$ are paraphrases and otherwise when $Y = 0$. We seek to separate the dependence between the distribution of positive and negative pairs, motivated by the observation of the presence of bias in the negative pairs. To begin, we model the distribution of sentences by splitting the sentence $s_2$ of length $n$ into the autoregressive product of individual words, where $w_2^{(i)}$ denotes the $i$th word in $s_2$. By applying Bayesian Inference Law, we have:

{\small
\begin{align*}
P(Y = y|s_1, s_2) &= \frac{P(Y = y|s_1)\Pi_{i=1}^{n}P(w_2^{(i)}|s_1, Y = y, w_2^{(1:i-1)})}{\Pi_{i=1}^{n}P(w_2^{(i)}|w_2^{(1:i-1)},s_1)}. \stepcounter{equation}\tag{\theequation}\label{probability decomposition}
\end{align*}
}

Subtracting the logarithm for $Y = 1$ and $Y = 0$, we get:

{\small
\begin{align*}
& \log(P(Y = 1|s_1,s_2)) - \log(P(Y = 0|s_1,s_2)) \\
&= (\log P(Y = 1|s_1)  - \log P(Y = 0 |s_1)) \\
&+ (\sum_{i=1}^n\log P(w_2^{(i)}|s_1, Y = 1, w_2^{(1:i-1)}) - (\sum_{i=1}^n\log P(w_2^{(i)}|s_1, Y = 0, w_2^{(1:i-1)}) )\\
&= (\log P(Y = 1)  - \log P(Y = 0 )) \\
&+ (\sum_{i=1}^n\log P(w_2^{(i)}|s_1, Y = 1, w_2^{(1:i-1)}) - (\sum_{i=1}^n\log P(w_2^{(i)}|s_1, Y = 0, w_2^{(1:i-1)}) ).
\stepcounter{equation}\tag{\theequation}\label{p equation}
\end{align*}
}

In this way, we break the probability inference into 3 terms resulting in Eqn.~\ref{p equation}: (1) $(\log P(Y = 1)  - \log P(Y = 0 ))$, which should just be a constant; (2) $(\sum_{i=1}^n\log P(w_2^{(i)}|s_1, Y = 1, w_2^{(1:i-1)})$, which depends only on the distribution of positive pairs; (3)$(\sum_{i=1}^n\log P(w_2^{(i)}|s_1, Y = 0, w_2^{(1:i-1)})$, which depends only on the distribution of negative pairs. We define the score of confidence as follows:

{\small
\begin{align*}
S(s_1, s_2) &= \underbrace{(\sum_{i=1}^n\log P(w_2^{(i)}|s_1, Y = 1, w_2^{(1:i-1)})}_\text{Positive Model} - \underbrace{(\sum_{i=1}^n\log P(w_2^{(i)}|s_1, Y = 0, w_2^{(1:i-1)}) )}_\text{Negative Model}.
\stepcounter{equation}\tag{\theequation}\label{final equation}
\end{align*}
}

In the above, we are now left with two terms, the first representing the positive model and the second the negative model. If we were to train the two terms together, the effects of the negative pairs in the resulting model can never be removed during inference, which we have observed to be a major source of bias. To avoid this, we propose to train the first term and the second term separately, and then subsequently ensemble them based on the degree that a given pair is out of distribution. We train each model on top of the pretrained autoregressive transformer described in \citep{DBLP:journals/corr/abs-1910-13461} known as BART. Given $s_1$ and $s_2$, we feed $s_1$ into the encoder as the condition, shift $s_2$ to the right by one-token, and feed shifted $s_2$ to the decoder. While the decoder proceeds autoregressively, we record the next-word probability distribution. We calculate the cross entropy between the next-word probability distribution and the target token in $s_2$ to update the model parameters. Note that here the Bayesian formulation has been similarly raised in some of the previous work like \citet{moore-lewis-2010-intelligent}, but to the best of our knowledge, we're the first to propose this Bayesian formulation to control the reliance on different components of the model.

\subsection{Ensembling} \label{ensembling}
To combine the positive and negative model, if we know a priori $\mathcal{D}^t$ is in the same distribution as $\mathcal{D}^s$, we can directly substitute the prediction of the positive and negative model into Eqn.~\ref{final equation}. We will refer to this as the In-distribution Predictor (IDP). If we have reason to believe that there is a significant distribution shift between $\mathcal{D}^s$ and $\mathcal{D}^t$ (e.g., different sources of corpus and different dataset collection procedure), we observe empirically that we should only utilize the positive model and disregard the negative model due to the bias it introduces. We will refer to this as the Out-of-distribution Predictor (OODP).

\subsubsection{Automatic Ensembling} \label{GAP description}
However, in most cases, we have little or no knowledge of the testing distribution, in which case we need to automatically decide how important the negative model is by detecting how much a test pair is in the same distribution as the training set. We adopt a weighted interpolation between a constant and the negative model in addition to the positive model as follow:

{\small
\begin{align*}
&S(s_1, s_2) = \log P(s_2|s_1, Y = 1) - (1 - \lambda(s_1,s_2)) \log P(s_2|s_1, Y = 0) - \lambda(s_1,s_2) C,
\stepcounter{equation}\tag{\theequation}\label{ood emsemble}
\end{align*}
}

where $\lambda(s_1,s_2)$ is a weight parameter depending on $s_1$ and $s_2$, and $C$ is a constant that achieves a regularization effect. See Appendix for ablations on how $C$ can be set. $P(s_2|s_1, Y=1)$ and $P(s_2|s_1, Y=0)$ are the same terms in Eqn.\ref{final equation}. To automatically assign $\lambda(s_1, s_2)$ for different sentence pairs, we measure an out-of-distribution score for $(s_1,s_2)$ with regard to the training distribution. Specifically, we use the same set of training data, comprising both positive and negative pairs, from $\mathcal{D}^s$, on which we train another autoregressive model, which we will refer to as the distribution model. The distribution model is trained by feeding an empty string into the encoder and the concatenated $s_1$ and $s_2$ into the decoder, with the training goal of predicting the next token. We measure the perplexity of each sentence pair $(s_1, s_2)$ using the distribution model based on the following formula, $w^i$ being the $i$th token of the concatenated $(s_1, s_2)$ of length $n$:

{\small
\begin{align*}
    PP(s_1, s_2) = \sqrt[n]{(\prod_{i=1}^{n}\frac{1}{P(w^i|w^{1:i-1})}}.\stepcounter{equation}\tag{\theequation}\label{perplexity}
\end{align*}
}

We then fit a Weibull distribution to the perplexity of a held-back set of validation data, so that it can better model the right-skewed property of the distribution. We derive the exponential parameter $a$, the shape parameter $c$, the location parameter $loc$, and the scale parameter $scale$. During testing, $\lambda(s_1, s_2)$ can now be determined as:

{\small
\begin{align*}
    \lambda(s_1, s_2) =  cdf(PP(s_1,s_2), Weibull(a, c, loc, scale)).
    \stepcounter{equation}\tag{\theequation}\label{lambda definition}
\end{align*}
}

For the final prediction, we predict the sentence pair to be paraphrase if $S(s_1, s_2) \ge 0$ and non-paraphrase otherwise. This forms what we referred to earlier as GAP (Generalized Autoregressive Paraphrase-Identification).

\subsubsection{Capturing Interplay Between Positive and Negative Pairs} \label{GAPX description}
In practice, training a positive and negative model separately disregards the interplay between the positive and negative pairs, which could be important when the test pairs are in-distribution. To capture such interplay, we utilize both positive and negative pairs to train a discriminative model for sequence classification. Specifically, we first define a thresholding function based on the value of $\lambda$:

{\small
\begin{align*}
\tau(\lambda)=
\begin{cases}
0& \text{$\lambda < 0.9$}\\
1& \text{$else.$}
\end{cases}
\stepcounter{equation}\tag{\theequation}\label{tau definition}
\end{align*}
}

We then ensemble the discriminative model and GAP using the value of $\tau(\lambda)$:

{\small
\begin{align*}
S^*(s_1, s_2) = M(1- \tau(\lambda(s_1, s_2)))(P(Y=1|s_1, s_2) - \frac{1}{2}) +  \tau(\lambda(s_1, s_2))S(s_1, s_2),
\stepcounter{equation}\tag{\theequation}\label{gapx equation}
\end{align*}
}

where $P(Y=1|s_1, s_2)$ can be estimated by any discriminative model, and $M$ is a sufficiently large constant. Note that this definition is essentially the same as trusting the discriminative model when we do not have statistical evidence that the pair is out-of-distribution (p-test < 10\%) while trusting the GAP model otherwise. For the final prediction, we predict the sentence pair to be paraphrase if $S*(s_1, s_2) \ge 0$ and non-paraphrase otherwise. This defines GAPX (Generalized Autoregressive Paraphrase Identfication X), for which we set $M$ to be sufficiently large ($> 1000$), so that when comparing the model confidence for different pairs, the score of the discriminative model will be prioritized.

\section{Experiments}
Our experiments are designed to (1) verify that the task of paraphrase identification suffers from biases in the datasets that is the main obstacle to generalization in this field of study, (2) test the accuracy of our perplexity based out-of-distribution detection method, and (3) test that balancing the utilization of the negative model can help outperform the state-of-the-art in the face of distribution shift, without losing in the in-distribution scenarios.

\subsection{Datasets}
\label{sec:datasets}
We compare our method against the other state-of-the-art methods on different combinations of the following datasets:

\begin{itemize}
    \item Quora Question Pair (QQP) consists of over 400,000 lines of potential question duplicate pairs. Since the original sampling method returns an imbalanced dataset, the authors attempt to balance it with additional negative pairs collected from similar topics to make them harder. Note that to scale QQP down to approximately the same size of PAWS and PIT (see below), we take the first 10k training pairs and 2k testing pairs from the train and test split by \citet{DBLP:journals/corr/WangHF17}.
    
    \item World Machine Translation Metrics Task 2017 (WMT) \citep{Bojar2017ResultsOT} contains in total 3793 manual ratings of machine translations from 7 languages to English. Each rating result contains a source sentence in the source language, a reference sentence in English (ground truth translation), a machine translated sentence in English, and a manual rating of the quality of translations. We take the ground-truth reference sentence and the machine translated sentence as the sentence pair. Sentences with a higher quality score ($>0$) are labeled as paraphrases, while those with lower quality scores ($\le 0$) are labeled otherwise. The resulting paraphase identification dataset is balanced. Note that this dataset is significantly smaller than other datasets, so we only use it as a test set.
    \item Paraphrase and Semantic Similarity in Twitter (PIT) \citep{Xu-EtAl-2014:TACL, xu2015semeval} contains 18762 sentence pairs automatically extracted from a similar distribution of topics as QQP. Annotators manually assigned integer scores from 0 to 5 to each sentence pair, representing the degree of similarity between the sentence pair. To make it a paraphrase identification dataset, we label sentence pairs with low scores (0, 1) as non-paraphrases and those with high scores (4,5) as paraphrases. The original dataset is unbalanced, so we randomly sample a maximum balanced subset of the dataset. The original test set processed in this way shrinks to only 350, and is not comparable to the other datasets. Hence, we use the original development data of size 1896 as the test set while keeping original test set of size 350 for development. The training set contains 5332 sentence pairs.
    
    \item Paraphrase Adversarials from Word Scrambling (PAWS) \citep{DBLP:journals/corr/abs-1904-01130} contains 49,401 sentence pairs, each of which is constructed from the same bag-of-word to make the evaluation more challenging. Most of the negative pairs are generated by word swapping while positive pairs are supplemented by back translation. This dataset contains paraphrase pairs that are the adversarial counterparts of standard paraphrase datasets such as QQP and PIT.
\end{itemize}

With these datasets, we perform experiments where different models (Sec.~\ref{sec:baselines}) are trained on one dataset and evaluated on another in order to observe whether their performance hold in the face of distribution shift.

\subsection{Benchmarks}
\label{sec:baselines}
We benchmark the paraphrase identification task with these methods: 

\begin{itemize}
    \item \textbf{BOW} \citep{DBLP:journals/corr/abs-1904-01130} represents two input sentences with bag of words. The bag of words representation of each input sentence is passed through a fully-connected network and cosine similarity between of the final layer is used to compute the classification output.  
    
    \item \textbf{BiLSTM} \cite{DBLP:journals/corr/abs-1806-04330} passes each of the two input sentences through a bidirectional LSTM network. The output state of the two sentences are then concatenated together and passed through a fully-connected network to get the classification output.

	\item \textbf{BERT}~\citep{DBLP:journals/corr/abs-1810-04805} is representative of the state-of-the-art transformer methods for text classification. We finetune the pretrained model "bert-base-uncased" in a standard way. We concatenate the sentence pair separated by a [SEP] token and take the [CLS] token as aggregate representation for the sentence pair. The embedding for the [CLS] token is then fed into an output layer for classification.
	
	\item \textbf{BART}~\citep{DBLP:journals/corr/abs-1910-13461} is the original transformer model that we build on by finetuning the pretrained model "bart-base-uncased". We concatenate the sentence pair separated by a </s> token, feeding it both into the encoder and the decoder. We use the <s> token at the end of the sentence pair for aggregate representation so that it can attend to decoder states from the complete input. The embedding for <s> is then fed into an output layer for classification.
	
	\item \textbf{RoBERTa}~\citep{DBLP:journals/corr/abs-1907-11692} shares the same transformer architecture with BERT, but uses a more robust pretraining strategy, and as a result performs better than BERT in many NLP tasks \citep{DBLP:journals/corr/abs-1907-11692}. For our experiments, we employ RoBERTa in the same way as BERT.
	
	\item \textbf{IDP} (In-distribution Predictor) is our model for in-distribution prediction if we know a priori that the testing pairs come from the same distribution as the training pairs. It combines the positive and negative models as given in Eqn.~\ref{final equation}. 
    
    \item \textbf{OODP} (Out-of-distribution Predictor) is our model for out-of-distribution prediction if we know a priori that the testing and training pairs are not in the same distribution. It only makes use of the positive model. We expect our OODP to have better generalizability than our IDP, because of its reduced reliance on negative examples. 
	
	\item \textbf{GAP} (Generalized Autoregressive Paraphrase-Identification) is our method that utilizes the perplexity based out-of-distribution detector to automatically control the reliance on the negative model, using Eqn.~\ref{ood emsemble}. This setting is different from IDP and OODP in that we do not have a priori knowledge of the test distribution.
	
    \item \textbf{GAPX} (Generalized Autoregressive Paraphrase-Identification X) ensembles GAP with RoBERTa described above, because we found RoBERTa to be the strongest baseline for in-distribution prediction. The intention is to capture via RoBERTa the interplay between positive and negative pairs. As depicted in Eqn.~\ref{gapx equation}, when we do not have significant evidence that the given pair is out of distribution (p-test > 10\%), we trust the prediction given by the discriminative model (RoBERTa). Otherwise, we trust GAP.
\end{itemize}

Following the previous literature in sentence matching \citep{DBLP:journals/corr/WangHF17, DBLP:journals/corr/abs-1810-04805}, we mainly use macro F1 score (F1) and accuracy score (ACC) to evaluate the models. Results based on Area-under-curve of the Receiver Operating Characteristic Curve (AUROC), a common metrics used to evaluate out-of-distribution metrics, are also provided in Appendix.\ref{auc results}.

\subsection{Measuring Distribution Shift} \label{measure distribution shift}
To situate our experiments properly, we note that different datasets does not equate to different distributions. It is thus important for us to be able to measure the distribution gap between datasets, and shed light on the models that perform the best when transferring between datasets with high distribution gap. Metrics that is per sample based such as our perplexity measure are not suitable for measuring at dataset level. To this end, we take a look at the Reverse Classification Accuracy* (RCA*) metric that has been proposed to predict the drop of model performance \citep{elsahar-galle-2019-annotate} and model selection \citep{Fan2006ReverseTA, inproceedings}. Here, given $D^s$ and $D^t$, we first train a model $M1$ from the training set of $D^s$. We then take a certain amount of samples from the target distribution $D^t$ (1000 in our experiments), and use $ M1$ to relabel them. The relabeled pairs are then utilized to train a new model $M2$. We measure the performance (in terms of AUC or ACC) of $M2$ on a held-out test set from $D^s$. As a control, we train another model $M3$ following the same procedure except the relabeled data comes from $D^s$. We denote the performance drop from $M3$ to $M2$ as the RCA* score indicative of the distribution gap.

To calibrate RCA*, for each dataset, we randomly selected 1000 held-out pairs and measure its shift from the dataset itself (which we expect to characterize an in-distribution RCA*). After 100 repetitions of measurements, we get a probability distribution of RCA* scores for each distribution in itself. Calibration results are reported in Appendix.~\ref{rca_distribution}. All the distributions share a mean of around 0 and a standard deviation around 2. By a p-test of 10\%, it is unlikely that datasets with a RCA* greater than 4.0 would belong to the same distribution. Based on this observation, most of the transfers between different datasets are likely out-of-distribution, except for transfers from QQP to PIT and vice versa, which have a RCA* score of 2.8 and 3.4 respectively. This is probably because QQP and PIT are curated in a similar fashion, where they extract sentences with similar topics from social platforms, while all other datasets adopt different strategies to collect their data.

Finally, it is important to note that although RCA* provides a good estimate of the distribution shift at the dataset level, it's utility does not easily extend to Eqn.~\ref{ood emsemble} and \ref{gapx equation} as opposed to per sample out-of-distribution metrics. RCA* assumes availability of the entire test set, while in real world, we are much more likely to get a single or small batches of test pairs at a time, all of which could even be from different distributions.

\subsection{Implementation details} \label{implementation details}
In practice, for testing, it helps to average the cross entropy by the length of the sentence, and average the cross entropy of generating $s_2$ from $s_1$ and generating $s_1$ from $s_2$. To optimize conditional sentence generators, we use Adam optimizer with learning rate 2e-5. We adopt cross entropy loss for each word logit. All experiments are run on Nvidia 2080 Ti with 11 GB memory.

\begin{figure}[h!]
    \centering
    \includegraphics[width=\textwidth]{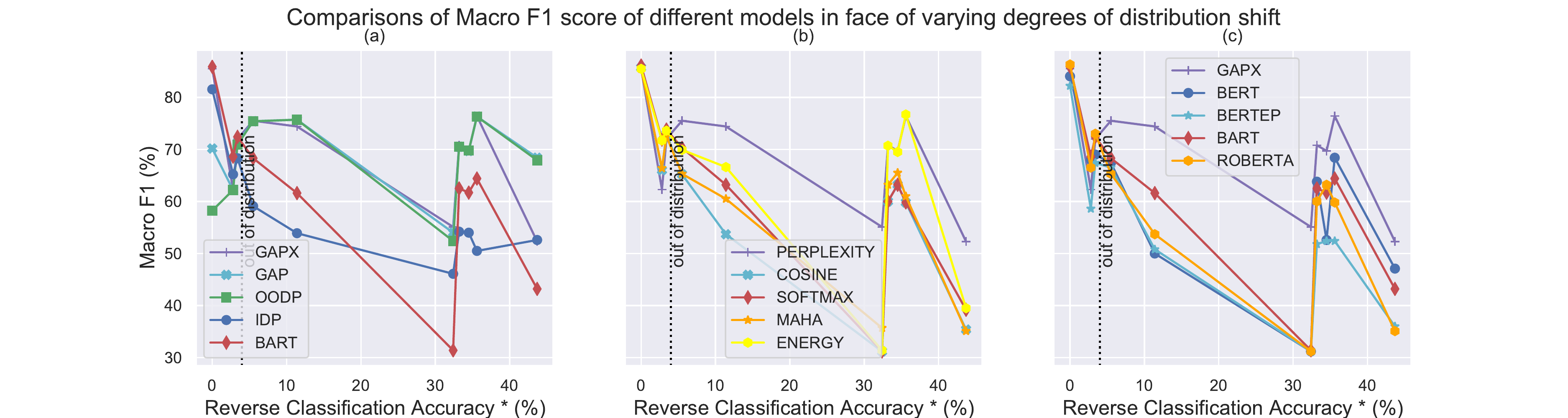}
    \caption{Comparing Macro F1 scores of different models at varying degrees of distribution shift.}
    \label{f1plots}
\end{figure}
\begin{figure}[h!]
    \centering
    \includegraphics[width=\textwidth]{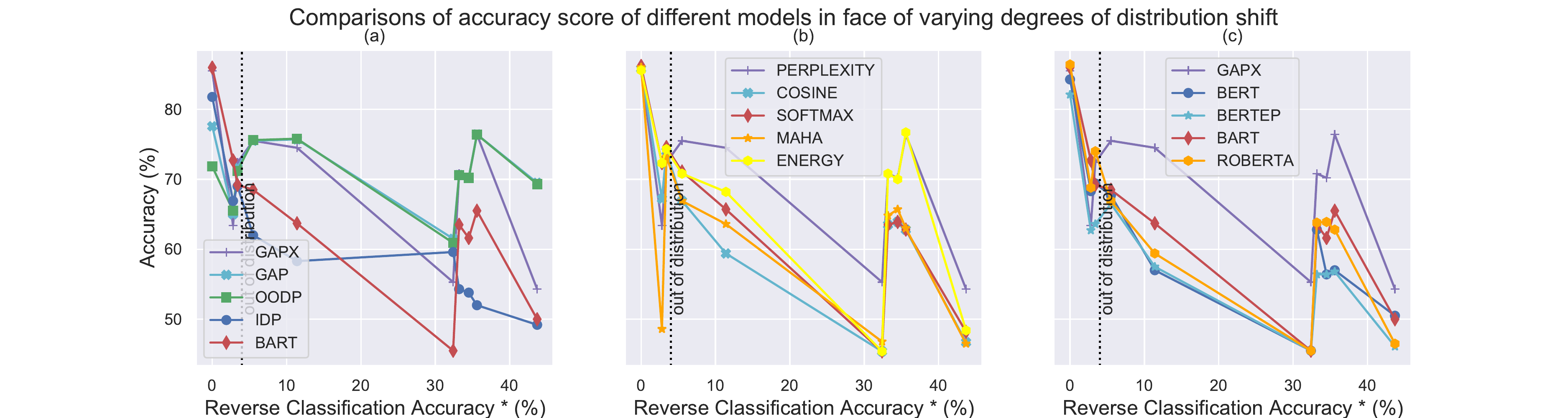}
    \caption{Comparing ACC score of different models at varying degrees of distribution shift.}
    \label{accplots}
\end{figure}

\subsection{Main Results}
\begin{table*}[!h]
\resizebox{\columnwidth}{!}{%
\begin{tabular}{lcccccccc}
Model  & QQP & PIT & PIT & PAWS & PAWS & PAWS & QQP& \multirow{2}{*}{average}\\
 & -> WMT   & -> WMT & -> PAWS & -> QQP & -> PIT & -> WMT  &-> PAWS \\
 & (5.5) & (11.4) & (32.4) & (33.2) & (34.5) & (35.6)  & (43.7)\\
 \hline
BOW & 34.6/51.5 & 33.3/51.4 & 35.3/54.7 & 33.3/50.0 & 34.3/50.2 & 34.8/51.7 & 35.3/54.7 & 34.4/52.0\\
BiLSTM & 34.4/51.5 & 50.7/51.1 & 48.6/48.7 & 36.8/50.4 & 43.3/50.6 & 34.9/51.2 & 37.1/54.7 & 40.8/51.2\\
BERT  & 67.4/67.7 & 50.0/57.7 & 31.2/45.5 &63.8 /62.8 & 52.6/56.4  & 68.4/57.0 & 47.1/50.5 & 54.4/56.8\\
BERT+EP  & 66.5/66.5 & 50.8/57.5 & 31.2/45.5 & 51.8/56.4 & 52.4/56.4  & 52.4/56.8  & 36.0/46.1 & 48.7/55.0\\
BART  & 68.3/68.5 & 61.6/63.7 & 31.4/45.5 & 62.5/63.5 & 61.7/61.6 & 64.4/65.5  & 43.2/50.0 & 56.2/59.8\\
RoBERTa & 65.3/66.9 & 53.7/59.4 & 31.2/45.5 & 60.0/63.6 & 63.2/63.9 & 59.8/62.8  & 35.1/46.5 & 52.6/58.4\\ 
\hline
IDP & 59.1/62.0 & 53.9/58.3 & 46.1/59.6 &54.2 /54.3 & 54.0/53.8 & 50.5/52.0 & 52.6/49.2 & 52.9/55.6\\
OODP   & 75.4/\textbf{75.6} & \textbf{75.7}/\textbf{75.8} & 52.4/60.9 & 70.5/70.6 & \textbf{69.8}/\textbf{70.2} & 76.3/\textbf{76.4}  & 67.9/69.3 & 69.7/\textbf{71.3}\\
GAP   & 75.4/75.5 & \textbf{75.7}/75.7  & 54.0/\textbf{61.5} & 70.5/70.6  & 69.7/\textbf{70.2} & 76.3/\textbf{76.4} & \textbf{68.4}/\textbf{69.5} & \textbf{70.0}/\textbf{71.3}\\
GAPX & \textbf{75.5}/75.5 & 74.4/74.5 &\textbf{ 55.1}/55.5 & \textbf{70.8}/\textbf{70.8} & 69.7/\textbf{70.2} & \textbf{76.4}/\textbf{76.4} & 52.3/54.3 & 67.7/68.2\\
\end{tabular}
}
\caption{Model performance on different out-of-distribution combinations of QQP, PAWS and PIT, in terms of macro F1/accuracy (ACC). Parenthesized is the RCA* score for each combination of datasets.}
\label{table1}
\end{table*}

\begin{table}[!h]
\centering
\begin{tabular}{lccccccr}
Model & QQP & PAWS  & PIT & QQP & PIT& \multirow{2}{*}{average}\\
& ->QQP & ->PAWS  & ->PIT & ->PIT & ->QQP\\
& (0) & (0)  & (0) & (2.8) & (3.4) \\
\hline
BOW & 51.3/57.8 & 48.8/56.8 & 33.3/50.0 & 41.7/49.7 & 33.3/50.0 & 40.7/52.9\\
BiLSTM & 61.6/63.6 & 43.6/53.0 & 50.6/51.1 & 51.7/52.9 & 41.1/49.0 & 49.7/53.9\\
BERT & 82.5/82.6 & 92.7/93.2 & 76.9/77.0& 68.0/68.3 & 69.0/69.4 & 77.8/78.1\\
BERT+EP & 81.6/81.7 & 89.7/89.7 & 75.3/74.9& 58.6/62.7 & 67.4/63.6 & 74.5/74.5\\
BART & 82.6/82.8 & 94.1/94.1 & 80.9/81.0& 68.6/72.7 & 72.4/69.2 & 79.7/80.0\\
RoBERTa & 84.4/84.5 & 93.5/93.6 & 81.0/81.1 &  66.5/68.8 & 73.0/74.9 & 79.7/80.6\\
\hline
OODP & 65.3/73.2 & 67.9/77.1 & 41.5/65.2& 62.2/65.5 & 65.2/71.2 & 60.4/70.4\\
IDP & 79.0/79.1 & 88.2/88.5 & 77.4/77.7& 65.2/66.9 & 68.3/69.0 & 75.6/76.2\\
GAP & 68.8/71.0 & 85.1/85.2 & 56.6/76.5 & 62.2/65.0 & 71.3/71.7 & 68.8/73.9\\
GAPX & 84.4/84.5 & 92.7/92.7& 79.3/79.3& 62.3/63.4 & 72.0/72.4 & 78.1/78.5\\
\end{tabular}
\caption{Model performance for in-distribution performances on QQP, PAWS, and PIT, in terms of macro F1/accuracy (ACC). Parenthesized is the RCA* score for each combination of datasets. }
\label{table2}
\end{table}


\paragraph{Bias in Negative Pairs}
To understand whether negative pairs are major sources of bias, we plot the Macro F1 and ACC score against RCA* in Fig.~\ref{f1plots}(a) and Fig.~\ref{accplots}(a), comparing the performances of OODP, IDP, and BART, all finetuned from the same pretrained checkpoint. The x-axis is plotted in ascending order of RCA* between pairs of datasets given in Table~\ref{table1} and \ref{table2}. There are three pairs that have RCA* score of 0 (Table~\ref{table2}), for which we average the performance in the plots. Both the F1 and ACC plots share a similar pattern. In the in-distribution region, where the RCA* score is less than 4\% (Fig.~\ref{rca_distribution}), BART and IDP achieves similarly high performances of 79.7\% and 75.6\% F1 on average respectively. The 4.1\% gap in F1 is potentially due to the fact that IDP trains a negative model and positive model separately, neglecting the interaction between positive and negative pairs. Comparatively, the performance of OODP is significantly inferior to the other two, with only 60.4\% average F1. This changes in the out-of-distribution region, where the RCA* score is now greater than 4\%. OODP turns out to be the leading model over BART and IDP. With increasing degree of distribution shift, we observe that both BART and IDP are especially fragile and their performance drop significantly. When the RCA* is greater than 20, both models' F1 drop to as low as around 60\%, which can be hardly useful in practice. In contrast, OODP maintains an advantage in F1 of as much as 10-20\% throughout the out-of-distribution region. Since the only difference between OODP and IDP is that OODP transfers only the positive model, it confirms our hypothesis that the negative model does not generalize as well as the positive model.

\paragraph{Importance of the Interplay between Positive and Negative Pairs}
To understand the necessity of capturing the interplay for test pairs that are in distribution, we compare the performances of GAP, GAPX, and BART in Fig.~\ref{f1plots}(a) and Fig.~\ref{accplots}(a). GAP only ensembles the positive and negative model trained separately, so does not contain any interplay information. On the other hand, both BART, trained with positive and negative pairs together, and GAPX (Sec.~\ref{GAPX description}) capture interplay information. As shown in the plot, the performance of GAP in the in-distribution region is not directly comparable to BART with a gap of 10.9\% in macro F1. In contrast, GAPX's in-distribution result has a much smaller margin of 1.6\% macro F1 compared to BART, yet, by automatically weighing the contribution of the discriminative model, GAPX also closely matches the performance of GAP in the out-of-distribution region with only a 2.3\% loss in macro F1.

\paragraph{Effectiveness of Perplexity-based Ensembling}
We also substitute $PP(s_1, s_2)$ with other state-of-the-art out-of-distribution metrics used for estimating the probability in Eqn.~\ref{lambda definition} for GAPX. Specifically, we compare our perplexity metric with Maximum Softmax Probability (SOFTMAX) \citep{DBLP:journals/corr/HendrycksG16c, DBLP:journals/corr/abs-2002-11297, Bergman2020ClassificationBasedAD}, Energy Score (ENERGY) \citep{DBLP:journals/corr/abs-2010-03759}, Mahalanobis Distance (MAHA) \citep{https://doi.org/10.48550/arxiv.1807.03888}, and COSINE \citep{DBLP:journals/corr/abs-2104-08812}. See Appendix for more details. The results are given in Fig.~\ref{accplots}(b) and Fig.~\ref{f1plots}(b). F1 and ACC are all similar in the in-distribution region. However, in the out-of-distribution region, SOFTMAX, MAHA, and COSINE start to perform poorly. ENERGY turned out to be the most robust but is still obviously not matching our perplexity metric, often by a large margin.

\paragraph{Generalization}
We implemented one of the most popular methods for domain generalization in paraphrase identification, Expert Product \citep{DBLP:journals/corr/abs-1909-03683}, as a potential strong baseline for handling dataset bias. As described in Sec.~\ref{sec:baselines}, we train a BERT classifier with only the first sentence $s_1$ as the biased model for BERT+EP. We report the results in Fig.~\ref{accplots}(c) and Fig.~\ref{aucplots}(c), together with BERT, BART and RoBERTa. In addition, we also provide performance of traditional methods like BoW~\cite{DBLP:journals/corr/abs-1904-01130} and ESIM~\cite{DBLP:journals/corr/ChenZLWJ16} in Table~\ref{table1} and Table~\ref{table2}. BERT, BART, and RoBERTa all produce similar results on all combinations of datasets. We observe that their performances are consistently better than traditional methods like BoW and ESIM, showing that pretraining and finetuning can indeed improve the generalizability of classifiers for paraphrase identification. However, their performances in out-of-distribution setting are still far from their in-distribution performances, with accuracy below 65\% in most of the cases (around 20\% drop). GAPX maintains an absolute margin of around 10\% in terms of ACC and an absolute margin of 7-20\% in terms of F1 in the out-of-distribution region. The best transformer-based models in the out-of-distribution region is BART, with an average of 56.2\% in F1 and 59.8\% in ACC, while GAPX maintains an average of 67.7\% in F1 and 68.2\% in ACC, with an absolute gain of 11.5\% in F1 and 8.4\% in ACC. What is also encouraging is that GAPX's performance is close to that of OODP, which is promising that we do not need \textit{a priori} information on the domain gap between the source and target. Lastly, BERT+EP fails to provide much gain, which we conjecture is due to the difficulty of ``finding'' the right bias model or features.


\section{Conclusion}
We have shown that negative samples introduce bias that prevent the generalization of paraphrase identification models. To overcome, we present a novel paradigm to train separate models for the distribution of positive and negative samples independently, and utilize a perplexity based out-of-distribution detection to ensemble them automatically. Experiments show that our method achieves an average of 11.5\% gain of F1 and 8.4\% gain of ACC in various different out-of-distribution scenarios over other state-of-the-art methods. 

\subsection{Limitations} \label{limitations}
Our methodology is specifically designed for only ``verification'' problems, where the samples come in pairs.
Scenarios that involve other types of bias will require non-trivial turn-key formulations to explicitly model the source of bias (much like the negative model). 




\section{Acknowledgements}
This work is sponsored by Meta AI. We would also like to thank Prof. Yoav Artzi for a helpful discussion in the early stage of this project.

\bibliography{anthology,custom}

\begin{thebibliography}{66}
\providecommand{\natexlab}[1]{#1}
\providecommand{\url}[1]{\texttt{#1}}
\expandafter\ifx\csname urlstyle\endcsname\relax
  \providecommand{\doi}[1]{doi: #1}\else
  \providecommand{\doi}{doi: \begingroup \urlstyle{rm}\Url}\fi

\bibitem[Abati et~al.(2019)Abati, Porrello, Calderara, and
  Cucchiara]{abati2019latent}
D.~Abati, A.~Porrello, S.~Calderara, and R.~Cucchiara.
\newblock Latent space autoregression for novelty detection.
\newblock In \emph{Proceedings of the IEEE/CVF Conference on Computer Vision
  and Pattern Recognition}, pages 481--490, 2019.

\bibitem[Agrawal et~al.(2018)Agrawal, Batra, Parikh, and
  Kembhavi]{agrawal2018don}
A.~Agrawal, D.~Batra, D.~Parikh, and A.~Kembhavi.
\newblock Don't just assume; look and answer: Overcoming priors for visual
  question answering.
\newblock In \emph{Proceedings of the IEEE Conference on Computer Vision and
  Pattern Recognition}, pages 4971--4980, 2018.

\bibitem[Amirkhani et~al.(2020)Amirkhani, AzariJafari, Pourjafari,
  Faridan-Jahromi, Kouhkan, and Amirak]{amirkhani2020farstail}
H.~Amirkhani, M.~AzariJafari, Z.~Pourjafari, S.~Faridan-Jahromi, Z.~Kouhkan,
  and A.~Amirak.
\newblock Farstail: A persian natural language inference dataset.
\newblock \emph{arXiv preprint arXiv:2009.08820}, 2020.

\bibitem[Anand et~al.(2018)Anand, Belilovsky, Kastner, Larochelle, and
  Courville]{DBLP:journals/corr/abs-1811-05013}
A.~Anand, E.~Belilovsky, K.~Kastner, H.~Larochelle, and A.~C. Courville.
\newblock Blindfold baselines for embodied {QA}.
\newblock \emph{CoRR}, abs/1811.05013, 2018.
\newblock URL \url{http://arxiv.org/abs/1811.05013}.

\bibitem[Bahng et~al.(2020)Bahng, Chun, Yun, Choo, and Oh]{pmlr-v119-bahng20a}
H.~Bahng, S.~Chun, S.~Yun, J.~Choo, and S.~J. Oh.
\newblock Learning de-biased representations with biased representations.
\newblock In H.~D. III and A.~Singh, editors, \emph{Proceedings of the 37th
  International Conference on Machine Learning}, volume 119 of
  \emph{Proceedings of Machine Learning Research}, pages 528--539. PMLR, 13--18
  Jul 2020.
\newblock URL \url{https://proceedings.mlr.press/v119/bahng20a.html}.

\bibitem[Bergman and Hoshen(2020)]{Bergman2020ClassificationBasedAD}
L.~Bergman and Y.~Hoshen.
\newblock Classification-based anomaly detection for general data.
\newblock \emph{ArXiv}, abs/2005.02359, 2020.

\bibitem[Bhagat and Hovy(2013)]{10.1162/COLI_a_00166}
R.~Bhagat and E.~Hovy.
\newblock {What Is a Paraphrase?}
\newblock \emph{Computational Linguistics}, 39\penalty0 (3):\penalty0 463--472,
  09 2013.
\newblock ISSN 0891-2017.
\newblock \doi{10.1162/COLI_a_00166}.
\newblock URL \url{https://doi.org/10.1162/COLI\_a\_00166}.

\bibitem[Bojar et~al.(2017)Bojar, Graham, and Kamran]{Bojar2017ResultsOT}
O.~Bojar, Y.~Graham, and A.~Kamran.
\newblock Results of the wmt17 metrics shared task.
\newblock In \emph{WMT}, 2017.

\bibitem[Bulusu et~al.(2020)Bulusu, Kailkhura, Li, Varshney, and Song]{9144212}
S.~Bulusu, B.~Kailkhura, B.~Li, P.~K. Varshney, and D.~Song.
\newblock Anomalous example detection in deep learning: A survey.
\newblock \emph{IEEE Access}, 8:\penalty0 132330--132347, 2020.
\newblock \doi{10.1109/ACCESS.2020.3010274}.

\bibitem[Cad{\`{e}}ne et~al.(2019)Cad{\`{e}}ne, Dancette, Ben{-}younes, Cord,
  and Parikh]{DBLP:journals/corr/abs-1906-10169}
R.~Cad{\`{e}}ne, C.~Dancette, H.~Ben{-}younes, M.~Cord, and D.~Parikh.
\newblock Rubi: Reducing unimodal biases in visual question answering.
\newblock \emph{CoRR}, abs/1906.10169, 2019.
\newblock URL \url{http://arxiv.org/abs/1906.10169}.

\bibitem[Cer et~al.(2017)Cer, Diab, Agirre, Lopez-Gazpio, and
  Specia]{cer-etal-2017-semeval}
D.~Cer, M.~Diab, E.~Agirre, I.~Lopez-Gazpio, and L.~Specia.
\newblock {S}em{E}val-2017 task 1: Semantic textual similarity multilingual and
  crosslingual focused evaluation.
\newblock In \emph{Proceedings of the 11th International Workshop on Semantic
  Evaluation ({S}em{E}val-2017)}, pages 1--14, Vancouver, Canada, Aug. 2017.
  Association for Computational Linguistics.
\newblock \doi{10.18653/v1/S17-2001}.
\newblock URL \url{https://aclanthology.org/S17-2001}.

\bibitem[Chen et~al.(2016)Chen, Zhu, Ling, Wei, and
  Jiang]{DBLP:journals/corr/ChenZLWJ16}
Q.~Chen, X.~Zhu, Z.~Ling, S.~Wei, and H.~Jiang.
\newblock Enhancing and combining sequential and tree {LSTM} for natural
  language inference.
\newblock \emph{CoRR}, abs/1609.06038, 2016.
\newblock URL \url{http://arxiv.org/abs/1609.06038}.

\bibitem[Chen et~al.(2020)Chen, Kornblith, Norouzi, and
  Hinton]{DBLP:journals/corr/abs-2002-05709}
T.~Chen, S.~Kornblith, M.~Norouzi, and G.~E. Hinton.
\newblock A simple framework for contrastive learning of visual
  representations.
\newblock \emph{CoRR}, abs/2002.05709, 2020.
\newblock URL \url{https://arxiv.org/abs/2002.05709}.

\bibitem[Choi et~al.(2019)Choi, Jang, and Alemi]{choi2019waic}
H.~Choi, E.~Jang, and A.~A. Alemi.
\newblock Waic, but why? generative ensembles for robust anomaly detection,
  2019.

\bibitem[Clark et~al.(2019)Clark, Yatskar, and
  Zettlemoyer]{DBLP:journals/corr/abs-1909-03683}
C.~Clark, M.~Yatskar, and L.~Zettlemoyer.
\newblock Don't take the easy way out: Ensemble based methods for avoiding
  known dataset biases.
\newblock \emph{CoRR}, abs/1909.03683, 2019.
\newblock URL \url{http://arxiv.org/abs/1909.03683}.

\bibitem[Devlin et~al.(2018)Devlin, Chang, Lee, and
  Toutanova]{DBLP:journals/corr/abs-1810-04805}
J.~Devlin, M.~Chang, K.~Lee, and K.~Toutanova.
\newblock {BERT:} pre-training of deep bidirectional transformers for language
  understanding.
\newblock \emph{CoRR}, abs/1810.04805, 2018.
\newblock URL \url{http://arxiv.org/abs/1810.04805}.

\bibitem[DeVries and Taylor(2018)]{https://doi.org/10.48550/arxiv.1802.04865}
T.~DeVries and G.~W. Taylor.
\newblock Learning confidence for out-of-distribution detection in neural
  networks, 2018.
\newblock URL \url{https://arxiv.org/abs/1802.04865}.

\bibitem[Elsahar and Gall{\'e}(2019)]{elsahar-galle-2019-annotate}
H.~Elsahar and M.~Gall{\'e}.
\newblock To annotate or not? predicting performance drop under domain shift.
\newblock In \emph{Proceedings of the 2019 Conference on Empirical Methods in
  Natural Language Processing and the 9th International Joint Conference on
  Natural Language Processing (EMNLP-IJCNLP)}, pages 2163--2173, Hong Kong,
  China, Nov. 2019. Association for Computational Linguistics.
\newblock \doi{10.18653/v1/D19-1222}.
\newblock URL \url{https://aclanthology.org/D19-1222}.

\bibitem[Fan and Davidson(2006)]{Fan2006ReverseTA}
W.~Fan and I.~Davidson.
\newblock Reverse testing: an efficient framework to select amongst classifiers
  under sample selection bias.
\newblock In \emph{KDD '06}, 2006.

\bibitem[Fort et~al.(2021)Fort, Ren, and
  Lakshminarayanan]{NEURIPS2021_3941c435}
S.~Fort, J.~Ren, and B.~Lakshminarayanan.
\newblock Exploring the limits of out-of-distribution detection.
\newblock In M.~Ranzato, A.~Beygelzimer, Y.~Dauphin, P.~Liang, and J.~W.
  Vaughan, editors, \emph{Advances in Neural Information Processing Systems},
  volume~34, pages 7068--7081. Curran Associates, Inc., 2021.
\newblock URL
  \url{https://proceedings.neurips.cc/paper/2021/file/3941c4358616274ac2436eacf67fae05-Paper.pdf}.

\bibitem[Grill et~al.(2020)Grill, Strub, Altch{\'{e}}, Tallec, Richemond,
  Buchatskaya, Doersch, Pires, Guo, Azar, Piot, Kavukcuoglu, Munos, and
  Valko]{DBLP:journals/corr/abs-2006-07733}
J.~Grill, F.~Strub, F.~Altch{\'{e}}, C.~Tallec, P.~H. Richemond,
  E.~Buchatskaya, C.~Doersch, B.~{\'{A}}. Pires, Z.~D. Guo, M.~G. Azar,
  B.~Piot, K.~Kavukcuoglu, R.~Munos, and M.~Valko.
\newblock Bootstrap your own latent: {A} new approach to self-supervised
  learning.
\newblock \emph{CoRR}, abs/2006.07733, 2020.
\newblock URL \url{https://arxiv.org/abs/2006.07733}.

\bibitem[He et~al.(2019{\natexlab{a}})He, Zha, and
  Wang]{DBLP:journals/corr/abs-1908-10763}
H.~He, S.~Zha, and H.~Wang.
\newblock Unlearn dataset bias in natural language inference by fitting the
  residual.
\newblock \emph{CoRR}, abs/1908.10763, 2019{\natexlab{a}}.
\newblock URL \url{http://arxiv.org/abs/1908.10763}.

\bibitem[He et~al.(2019{\natexlab{b}})He, Zha, and Wang]{he2019unlearn}
H.~He, S.~Zha, and H.~Wang.
\newblock Unlearn dataset bias in natural language inference by fitting the
  residual.
\newblock \emph{arXiv preprint arXiv:1908.10763}, 2019{\natexlab{b}}.

\bibitem[Hendrycks and Gimpel(2016)]{DBLP:journals/corr/HendrycksG16c}
D.~Hendrycks and K.~Gimpel.
\newblock A baseline for detecting misclassified and out-of-distribution
  examples in neural networks.
\newblock \emph{CoRR}, abs/1610.02136, 2016.
\newblock URL \url{http://arxiv.org/abs/1610.02136}.

\bibitem[Hendrycks et~al.(2018)Hendrycks, Mazeika, and
  Dietterich]{DBLP:journals/corr/abs-1812-04606}
D.~Hendrycks, M.~Mazeika, and T.~G. Dietterich.
\newblock Deep anomaly detection with outlier exposure.
\newblock \emph{CoRR}, abs/1812.04606, 2018.
\newblock URL \url{http://arxiv.org/abs/1812.04606}.

\bibitem[Hsu et~al.(2020)Hsu, Shen, Jin, and
  Kira]{DBLP:journals/corr/abs-2002-11297}
Y.~Hsu, Y.~Shen, H.~Jin, and Z.~Kira.
\newblock Generalized {ODIN:} detecting out-of-distribution image without
  learning from out-of-distribution data.
\newblock \emph{CoRR}, abs/2002.11297, 2020.
\newblock URL \url{https://arxiv.org/abs/2002.11297}.

\bibitem[Kozareva and Montoyo(2006)]{kozareva2006paraphrase}
Z.~Kozareva and A.~Montoyo.
\newblock Paraphrase identification on the basis of supervised machine learning
  techniques.
\newblock In \emph{International conference on natural language processing (in
  Finland)}, pages 524--533. Springer, 2006.

\bibitem[Lan and Xu(2018)]{DBLP:journals/corr/abs-1806-04330}
W.~Lan and W.~Xu.
\newblock Neural network models for paraphrase identification, semantic textual
  similarity, natural language inference, and question answering.
\newblock \emph{CoRR}, abs/1806.04330, 2018.
\newblock URL \url{http://arxiv.org/abs/1806.04330}.

\bibitem[Lee et~al.(2018)Lee, Lee, Lee, and
  Shin]{https://doi.org/10.48550/arxiv.1807.03888}
K.~Lee, K.~Lee, H.~Lee, and J.~Shin.
\newblock A simple unified framework for detecting out-of-distribution samples
  and adversarial attacks, 2018.
\newblock URL \url{https://arxiv.org/abs/1807.03888}.

\bibitem[Lewis et~al.(2019)Lewis, Liu, Goyal, Ghazvininejad, Mohamed, Levy,
  Stoyanov, and Zettlemoyer]{DBLP:journals/corr/abs-1910-13461}
M.~Lewis, Y.~Liu, N.~Goyal, M.~Ghazvininejad, A.~Mohamed, O.~Levy, V.~Stoyanov,
  and L.~Zettlemoyer.
\newblock {BART:} denoising sequence-to-sequence pre-training for natural
  language generation, translation, and comprehension.
\newblock \emph{CoRR}, abs/1910.13461, 2019.
\newblock URL \url{http://arxiv.org/abs/1910.13461}.

\bibitem[Li et~al.(2021)Li, Liu, Wang, and Wang]{9414944}
B.~Li, T.~Liu, B.~Wang, and L.~Wang.
\newblock Enhancing deep paraphrase identification via leveraging word
  alignment information.
\newblock In \emph{ICASSP 2021 - 2021 IEEE International Conference on
  Acoustics, Speech and Signal Processing (ICASSP)}, pages 7843--7847, 2021.
\newblock \doi{10.1109/ICASSP39728.2021.9414944}.

\bibitem[Liang et~al.(2017)Liang, Li, and
  Srikant]{DBLP:journals/corr/LiangLS17}
S.~Liang, Y.~Li, and R.~Srikant.
\newblock Principled detection of out-of-distribution examples in neural
  networks.
\newblock \emph{CoRR}, abs/1706.02690, 2017.
\newblock URL \url{http://arxiv.org/abs/1706.02690}.

\bibitem[Liu et~al.(2020)Liu, Wang, Owens, and
  Li]{DBLP:journals/corr/abs-2010-03759}
W.~Liu, X.~Wang, J.~D. Owens, and Y.~Li.
\newblock Energy-based out-of-distribution detection.
\newblock \emph{CoRR}, abs/2010.03759, 2020.
\newblock URL \url{https://arxiv.org/abs/2010.03759}.

\bibitem[Liu et~al.(2019)Liu, Ott, Goyal, Du, Joshi, Chen, Levy, Lewis,
  Zettlemoyer, and Stoyanov]{DBLP:journals/corr/abs-1907-11692}
Y.~Liu, M.~Ott, N.~Goyal, J.~Du, M.~Joshi, D.~Chen, O.~Levy, M.~Lewis,
  L.~Zettlemoyer, and V.~Stoyanov.
\newblock Roberta: {A} robustly optimized {BERT} pretraining approach.
\newblock \emph{CoRR}, abs/1907.11692, 2019.
\newblock URL \url{http://arxiv.org/abs/1907.11692}.

\bibitem[Marsi and Krahmer(2005)]{explorations-in-sentence}
E.~Marsi and E.~Krahmer.
\newblock Explorations in sentence fusion.
\newblock 01 2005.

\bibitem[Min et~al.(2019)Min, Wallace, Singh, Gardner, Hajishirzi, and
  Zettlemoyer]{min-etal-2019-compositional}
S.~Min, E.~Wallace, S.~Singh, M.~Gardner, H.~Hajishirzi, and L.~Zettlemoyer.
\newblock Compositional questions do not necessitate multi-hop reasoning.
\newblock In \emph{Proceedings of the 57th Annual Meeting of the Association
  for Computational Linguistics}, pages 4249--4257, Florence, Italy, July 2019.
  Association for Computational Linguistics.
\newblock \doi{10.18653/v1/P19-1416}.
\newblock URL \url{https://aclanthology.org/P19-1416}.

\bibitem[Moore and Lewis(2010)]{moore-lewis-2010-intelligent}
R.~C. Moore and W.~Lewis.
\newblock Intelligent selection of language model training data.
\newblock In \emph{Proceedings of the {ACL} 2010 Conference Short Papers},
  pages 220--224, Uppsala, Sweden, July 2010. Association for Computational
  Linguistics.
\newblock URL \url{https://aclanthology.org/P10-2041}.

\bibitem[Pidhorskyi et~al.(2018)Pidhorskyi, Almohsen, and
  Doretto]{NEURIPS2018_5421e013}
S.~Pidhorskyi, R.~Almohsen, and G.~Doretto.
\newblock Generative probabilistic novelty detection with adversarial
  autoencoders.
\newblock In S.~Bengio, H.~Wallach, H.~Larochelle, K.~Grauman, N.~Cesa-Bianchi,
  and R.~Garnett, editors, \emph{Advances in Neural Information Processing
  Systems}, volume~31. Curran Associates, Inc., 2018.
\newblock URL
  \url{https://proceedings.neurips.cc/paper/2018/file/5421e013565f7f1afa0cfe8ad87a99ab-Paper.pdf}.

\bibitem[Radford et~al.(2019)Radford, Wu, Child, Luan, Amodei, Sutskever,
  et~al.]{radford2019language}
A.~Radford, J.~Wu, R.~Child, D.~Luan, D.~Amodei, I.~Sutskever, et~al.
\newblock Language models are unsupervised multitask learners.
\newblock \emph{OpenAI blog}, 1\penalty0 (8):\penalty0 9, 2019.

\bibitem[Rei et~al.(2020{\natexlab{a}})Rei, Stewart, Farinha, and
  Lavie]{DBLP:journals/corr/abs-2009-09025}
R.~Rei, C.~Stewart, A.~C. Farinha, and A.~Lavie.
\newblock {COMET:} {A} neural framework for {MT} evaluation.
\newblock \emph{CoRR}, abs/2009.09025, 2020{\natexlab{a}}.
\newblock URL \url{https://arxiv.org/abs/2009.09025}.

\bibitem[Rei et~al.(2020{\natexlab{b}})Rei, Stewart, Farinha, and
  Lavie]{rei-etal-2020-comet}
R.~Rei, C.~Stewart, A.~C. Farinha, and A.~Lavie.
\newblock {COMET}: A neural framework for {MT} evaluation.
\newblock In \emph{Proceedings of the 2020 Conference on Empirical Methods in
  Natural Language Processing (EMNLP)}, pages 2685--2702, Online, Nov.
  2020{\natexlab{b}}. Association for Computational Linguistics.
\newblock \doi{10.18653/v1/2020.emnlp-main.213}.
\newblock URL \url{https://aclanthology.org/2020.emnlp-main.213}.

\bibitem[Reimers and Gurevych(2019)]{DBLP:journals/corr/abs-1908-10084}
N.~Reimers and I.~Gurevych.
\newblock Sentence-bert: Sentence embeddings using siamese bert-networks.
\newblock \emph{CoRR}, abs/1908.10084, 2019.
\newblock URL \url{http://arxiv.org/abs/1908.10084}.

\bibitem[Ren et~al.(2019)Ren, Liu, Fertig, Snoek, Poplin, Depristo, Dillon, and
  Lakshminarayanan]{NEURIPS2019_1e795968}
J.~Ren, P.~J. Liu, E.~Fertig, J.~Snoek, R.~Poplin, M.~Depristo, J.~Dillon, and
  B.~Lakshminarayanan.
\newblock Likelihood ratios for out-of-distribution detection.
\newblock In H.~Wallach, H.~Larochelle, A.~Beygelzimer, F.~d\textquotesingle
  Alch\'{e}-Buc, E.~Fox, and R.~Garnett, editors, \emph{Advances in Neural
  Information Processing Systems}, volume~32. Curran Associates, Inc., 2019.
\newblock URL
  \url{https://proceedings.neurips.cc/paper/2019/file/1e79596878b2320cac26dd792a6c51c9-Paper.pdf}.

\bibitem[Sellam et~al.(2020)Sellam, Das, and
  Parikh]{DBLP:journals/corr/abs-2004-04696}
T.~Sellam, D.~Das, and A.~P. Parikh.
\newblock {BLEURT:} learning robust metrics for text generation.
\newblock \emph{CoRR}, abs/2004.04696, 2020.
\newblock URL \url{https://arxiv.org/abs/2004.04696}.

\bibitem[Shen and Lam(2021)]{shen-lam-2021-towards}
X.~Shen and W.~Lam.
\newblock Towards domain-generalizable paraphrase identification by avoiding
  the shortcut learning.
\newblock In \emph{Proceedings of the International Conference on Recent
  Advances in Natural Language Processing (RANLP 2021)}, pages 1318--1325, Held
  Online, Sept. 2021. INCOMA Ltd.
\newblock URL \url{https://aclanthology.org/2021.ranlp-main.148}.

\bibitem[Thompson and Post(2020)]{DBLP:journals/corr/abs-2004-14564}
B.~Thompson and M.~Post.
\newblock Automatic machine translation evaluation in many languages via
  zero-shot paraphrasing.
\newblock \emph{CoRR}, abs/2004.14564, 2020.
\newblock URL \url{https://arxiv.org/abs/2004.14564}.

\bibitem[Tomar et~al.(2017)Tomar, Duque, T{\"a}ckstr{\"o}m, Uszkoreit, and
  Das]{tomar2017neural}
G.~S. Tomar, T.~Duque, O.~T{\"a}ckstr{\"o}m, J.~Uszkoreit, and D.~Das.
\newblock Neural paraphrase identification of questions with noisy pretraining.
\newblock \emph{arXiv preprint arXiv:1704.04565}, 2017.

\bibitem[Utama et~al.(2020)Utama, Moosavi, and
  Gurevych]{DBLP:journals/corr/abs-2009-12303}
P.~A. Utama, N.~S. Moosavi, and I.~Gurevych.
\newblock Towards debiasing {NLU} models from unknown biases.
\newblock \emph{CoRR}, abs/2009.12303, 2020.
\newblock URL \url{https://arxiv.org/abs/2009.12303}.

\bibitem[Vaswani et~al.(2017)Vaswani, Shazeer, Parmar, Uszkoreit, Jones, Gomez,
  Kaiser, and Polosukhin]{DBLP:journals/corr/VaswaniSPUJGKP17}
A.~Vaswani, N.~Shazeer, N.~Parmar, J.~Uszkoreit, L.~Jones, A.~N. Gomez,
  L.~Kaiser, and I.~Polosukhin.
\newblock Attention is all you need.
\newblock \emph{CoRR}, abs/1706.03762, 2017.
\newblock URL \url{http://arxiv.org/abs/1706.03762}.

\bibitem[Wang et~al.(2021)Wang, Fang, Khabsa, Mao, and
  Ma]{DBLP:journals/corr/abs-2104-14690}
S.~Wang, H.~Fang, M.~Khabsa, H.~Mao, and H.~Ma.
\newblock Entailment as few-shot learner.
\newblock \emph{CoRR}, abs/2104.14690, 2021.
\newblock URL \url{https://arxiv.org/abs/2104.14690}.

\bibitem[Wang et~al.(2016)Wang, Mi, and
  Ittycheriah]{DBLP:journals/corr/WangMI16a}
Z.~Wang, H.~Mi, and A.~Ittycheriah.
\newblock Sentence similarity learning by lexical decomposition and
  composition.
\newblock \emph{CoRR}, abs/1602.07019, 2016.
\newblock URL \url{http://arxiv.org/abs/1602.07019}.

\bibitem[Wang et~al.(2017)Wang, Hamza, and
  Florian]{DBLP:journals/corr/WangHF17}
Z.~Wang, W.~Hamza, and R.~Florian.
\newblock Bilateral multi-perspective matching for natural language sentences.
\newblock \emph{CoRR}, abs/1702.03814, 2017.
\newblock URL \url{http://arxiv.org/abs/1702.03814}.

\bibitem[Xu et~al.(2014)Xu, Ritter, Callison-Burch, Dolan, and
  Ji]{Xu-EtAl-2014:TACL}
W.~Xu, A.~Ritter, C.~Callison-Burch, W.~B. Dolan, and Y.~Ji.
\newblock Extracting lexically divergent paraphrases from {Twitter}.
\newblock \emph{Transactions of the Association for Computational Linguistics},
  2:\penalty0 435--448, 2014.
\newblock \doi{10.1162/tacl_a_00194}.
\newblock URL \url{https://www.aclweb.org/anthology/Q14-1034}.

\bibitem[Xu et~al.(2015)Xu, Callison-Burch, and Dolan]{xu2015semeval}
W.~Xu, C.~Callison-Burch, and B.~Dolan.
\newblock {SemEval-2015 Task} 1: Paraphrase and semantic similarity in
  {Twitter} ({PIT}).
\newblock In \emph{Proceedings of the 9th International Workshop on Semantic
  Evaluation ({S}em{E}val 2015)}, pages 1--11. Association for Computational
  Linguistics, 2015.
\newblock \doi{10.18653/v1/S15-2001}.
\newblock URL \url{https://www.aclweb.org/anthology/S15-2001}.

\bibitem[Yang et~al.(2021)Yang, Zhou, Li, and
  Liu]{DBLP:journals/corr/abs-2110-11334}
J.~Yang, K.~Zhou, Y.~Li, and Z.~Liu.
\newblock Generalized out-of-distribution detection: {A} survey.
\newblock \emph{CoRR}, abs/2110.11334, 2021.
\newblock URL \url{https://arxiv.org/abs/2110.11334}.

\bibitem[Yang et~al.(2019)Yang, Zhang, Tar, and
  Baldridge]{DBLP:journals/corr/abs-1908-11828}
Y.~Yang, Y.~Zhang, C.~Tar, and J.~Baldridge.
\newblock {PAWS-X:} {A} cross-lingual adversarial dataset for paraphrase
  identification.
\newblock \emph{CoRR}, abs/1908.11828, 2019.
\newblock URL \url{http://arxiv.org/abs/1908.11828}.

\bibitem[Yin and Sch{\"u}tze(2015)]{yin2015convolutional}
W.~Yin and H.~Sch{\"u}tze.
\newblock Convolutional neural network for paraphrase identification.
\newblock In \emph{Proceedings of the 2015 Conference of the North American
  Chapter of the Association for Computational Linguistics: Human Language
  Technologies}, pages 901--911, 2015.

\bibitem[Yuan et~al.(2021)Yuan, Neubig, and
  Liu]{DBLP:journals/corr/abs-2106-11520}
W.~Yuan, G.~Neubig, and P.~Liu.
\newblock Bartscore: Evaluating generated text as text generation.
\newblock \emph{CoRR}, abs/2106.11520, 2021.
\newblock URL \url{https://arxiv.org/abs/2106.11520}.

\bibitem[Zbontar et~al.(2021)Zbontar, Jing, Misra, LeCun, and
  Deny]{DBLP:journals/corr/abs-2103-03230}
J.~Zbontar, L.~Jing, I.~Misra, Y.~LeCun, and S.~Deny.
\newblock Barlow twins: Self-supervised learning via redundancy reduction.
\newblock \emph{CoRR}, abs/2103.03230, 2021.
\newblock URL \url{https://arxiv.org/abs/2103.03230}.

\bibitem[Zhai et~al.(2016)Zhai, Cheng, Lu, and
  Zhang]{DBLP:journals/corr/ZhaiCLZ16}
S.~Zhai, Y.~Cheng, W.~Lu, and Z.~Zhang.
\newblock Deep structured energy based models for anomaly detection.
\newblock \emph{CoRR}, abs/1605.07717, 2016.
\newblock URL \url{http://arxiv.org/abs/1605.07717}.

\bibitem[Zhang et~al.(2019{\natexlab{a}})Zhang, Kishore, Wu, Weinberger, and
  Artzi]{DBLP:journals/corr/abs-1904-09675}
T.~Zhang, V.~Kishore, F.~Wu, K.~Q. Weinberger, and Y.~Artzi.
\newblock Bertscore: Evaluating text generation with {BERT}.
\newblock \emph{CoRR}, abs/1904.09675, 2019{\natexlab{a}}.
\newblock URL \url{http://arxiv.org/abs/1904.09675}.

\bibitem[Zhang et~al.(2019{\natexlab{b}})Zhang, Baldridge, and
  He]{DBLP:journals/corr/abs-1904-01130}
Y.~Zhang, J.~Baldridge, and L.~He.
\newblock {PAWS:} paraphrase adversaries from word scrambling.
\newblock \emph{CoRR}, abs/1904.01130, 2019{\natexlab{b}}.
\newblock URL \url{http://arxiv.org/abs/1904.01130}.

\bibitem[Zhao et~al.(2019)Zhao, Peyrard, Liu, Gao, Meyer, and
  Eger]{DBLP:journals/corr/abs-1909-02622}
W.~Zhao, M.~Peyrard, F.~Liu, Y.~Gao, C.~M. Meyer, and S.~Eger.
\newblock Moverscore: Text generation evaluating with contextualized embeddings
  and earth mover distance.
\newblock \emph{CoRR}, abs/1909.02622, 2019.
\newblock URL \url{http://arxiv.org/abs/1909.02622}.

\bibitem[Zhong et~al.(2010)Zhong, Fan, Yang, Verscheure, and
  Ren]{inproceedings}
E.~Zhong, W.~Fan, Q.~Yang, O.~Verscheure, and J.~Ren.
\newblock Cross validation framework to choose amongst models and datasets for
  transfer learning.
\newblock pages 547--562, 09 2010.
\newblock ISBN 978-3-642-15938-1.
\newblock \doi{10.1007/978-3-642-15939-8_35}.

\bibitem[Zhou and Chen(2021)]{DBLP:journals/corr/abs-2104-08812}
W.~Zhou and M.~Chen.
\newblock Contrastive out-of-distribution detection for pretrained
  transformers.
\newblock \emph{CoRR}, abs/2104.08812, 2021.
\newblock URL \url{https://arxiv.org/abs/2104.08812}.

\bibitem[Zong et~al.(2018)Zong, Song, Min, Cheng, Lumezanu, Cho, and
  Chen]{zong2018deep}
B.~Zong, Q.~Song, M.~R. Min, W.~Cheng, C.~Lumezanu, D.~Cho, and H.~Chen.
\newblock Deep autoencoding gaussian mixture model for unsupervised anomaly
  detection.
\newblock In \emph{International conference on learning representations}, 2018.

\end{thebibliography}
\bibliographystyle{abbrvnat}




\section*{Checklist}

\begin{enumerate}

\item For all authors...
\begin{enumerate}
  \item Do the main claims made in the abstract and introduction accurately reflect the paper's contributions and scope?
    \answerYes{}. See our listed contributions in the introduction.
  \item Did you describe the limitations of your work?
    \answerYes{}. See the end of our conclusion, Sec.~\ref{limitations}.
  \item Did you discuss any potential negative societal impacts of your work?
    \answerNA{}. This is a traditional Natural Language Processing research topic. We do not see any conceivable negative social impact.
  \item Have you read the ethics review guidelines and ensured that your paper conforms to them?
    \answerYes{}
\end{enumerate}

\item If you are including theoretical results...
\begin{enumerate}
  \item Did you state the full set of assumptions of all theoretical results?
    \answerYes{}. See Sec.~\ref{separation}
        \item Did you include complete proofs of all theoretical results?
    \answerYes{}. See Section.~\ref{separation}
\end{enumerate}

\item If you ran experiments...
\begin{enumerate}
  \item Did you include the code, data, and instructions needed to reproduce the main experimental results (either in the supplemental material or as a URL)?
    \answerYes{}.
  \item Did you specify all the training details (e.g., data splits, hyperparameters, how they were chosen)?
    \answerYes{}. See Sec.\ref{implementation details}
        \item Did you report error bars (e.g., with respect to the random seed after running experiments multiple times)?
    \answerNA{} The baseline models in NLP that we are comparing to do not report error bars in their paper.
        \item Did you include the total amount of compute and the type of resources used (e.g., type of GPUs, internal cluster, or cloud provider)?
    \answerYes{}. See Sec.~\ref{implementation details}.
\end{enumerate}

\item If you are using existing assets (e.g., code, data, models) or curating/releasing new assets...
\begin{enumerate}
  \item If your work uses existing assets, did you cite the creators?
    \answerYes{}. See Appendix~\ref{sec:datasets}
  \item Did you mention the license of the assets?
    \answerNA{}. Those are all public datasets.
  \item Did you include any new assets either in the supplemental material or as a URL?
    \answerNA{}
  \item Did you discuss whether and how consent was obtained from people whose data you're using/curating?
    \answerNA{}. Those are all public datasets, allowing researching purpose.
  \item Did you discuss whether the data you are using/curating contains personally identifiable information or offensive content?
    \answerNA{}. Those data does not contain personal information
\end{enumerate}

\item If you used crowdsourcing or conducted research with human subjects...
\begin{enumerate}
  \item Did you include the full text of instructions given to participants and screenshots, if applicable?
    \answerNA{}
  \item Did you describe any potential participant risks, with links to Institutional Review Board (IRB) approvals, if applicable?
    \answerNA{}
  \item Did you include the estimated hourly wage paid to participants and the total amount spent on participant compensation?
    \answerNA{}
\end{enumerate}

\end{enumerate}


\newpage
\appendix

\section{AUROC of Main Results} \label{auc results}
We include full results of AUROC scores here.
\begin{table*}[!h]
\centering
\resizebox{\columnwidth}{!}{%
\begin{tabular}{lcccccccc}
Model  & QQP & PIT & PIT & PAWS & PAWS & PAWS & QQP & \multirow{2}{*}{average}\\
 & -> WMT   & -> WMT & -> PAWS & -> QQP & -> PIT & -> WMT  &-> PAWS & \\
 & (5.5) & (11.4) & (32.4) & (33.2) & (34.5) & (35.6)  & (43.7)\\
 \hline
BOW & 51.9 & 52.5 & 49.0 & 56.1 & 44.0 & 53.2 & 49.1 & 50.8\\
BiLSTM & 50.1 & 51.5 & 49.3 & 50.4 & 50.6 & 49.8 & 50.1 & 50.3\\
BERT  & 75.0 & 70.7 & 51.2 & 69.2 & 58.2  & 70.7 & 54.7 & 64.2\\
BERT+EP  & 73.5 & 74.6 & 52.2 &66.9 & 62.3  & 69.4  & 53.4 & 64.6\\
BART  & 75.7 & 76.4 & 53.3 & 71.7 & 65.0 & 77.6  & 56.6 & 68.0\\
RoBERTa & 77.9 & 76.9 & 52.4 & 77.4 & 71.2 & 80.6  & 54.3 & 70.1\\ 
\hline
IDP & 71.0 & 65.7 & 63.9 & 56.0 & 55.0 & 50.3 & 67.0 & 61.3\\
OODP   & 85.0 & \textbf{85.0} & \textbf{67.2} & 76.7 & \textbf{77.5} & \textbf{84.9}  & 74.4 & \textbf{78.7}\\
GAP   & \textbf{85.1} & 84.7 & 66.4 & 76.2 & \textbf{77.5} & \textbf{84.9} & \textbf{74.7} & 78.5\\
GAPX & 83.8 & 81.1 & 58.3 & \textbf{77.7} & 77.5 & 84.6 & 59.5 & 74.6\\
\end{tabular}
}
\caption{Model performance on different out-of-distribution combinations of QQP, PAWS and PIT, in terms of area under curve (AUROC).}
\label{table3}
\end{table*}

\begin{table}[!h]
\centering
\begin{tabular}{lcccccr}
Model & QQP & PAWS  & PIT & QQP & PIT& \multirow{2}{*}{average}\\
& ->QQP & ->PAWS  & ->PIT & ->PIT & ->QQP\\
& (0) & (0)  & (0) & (2.8) & (3.4) \\
\hline
BOW & 60.9 & 57.7 & 47.7 & 47.4 & 61.3 & 55.0\\
BiLSTM & 64.6 & 49.6 & 50.9 & 52.8 & 49.0 & 53.4 \\
BERT & 90.0 & 97.6 & 85.1& 77.3 &73.4 & 84.7\\
BERT+EP & 89.8 & 95.8 & 82.7 &76.1& 73.7 & 83.6\\
BART & 90.6 & 98.2 & 89.1& 78.2 & 77.9 & 86.8\\
RoBERTa & 92.0 & 98.7 & 90.2 &  82.0 & 80.0 & 88.6\\
\hline
OODP & 79.6 & 84.6 & 73.6&72.8 &76.4 & 77.4\\
IDP & 88.1 & 95.4 & 86.8 &77.9 & 74.6 & 84.6\\
GAP & 78.9 & 92.3 & 84.5 & 72.3 & 76.4 & 80.9\\
GAPX &90.7 & 98.1 & 87.4 & 72.5& 79.2 & 85.6\\
\end{tabular}
\caption{Model performance for in-distribution performances on QQP, PAWS, and PIT, in terms of area under curve (AUROC). }
\label{table4}
\end{table}

\begin{figure}[h!]
    \centering
    \includegraphics[width=\textwidth]{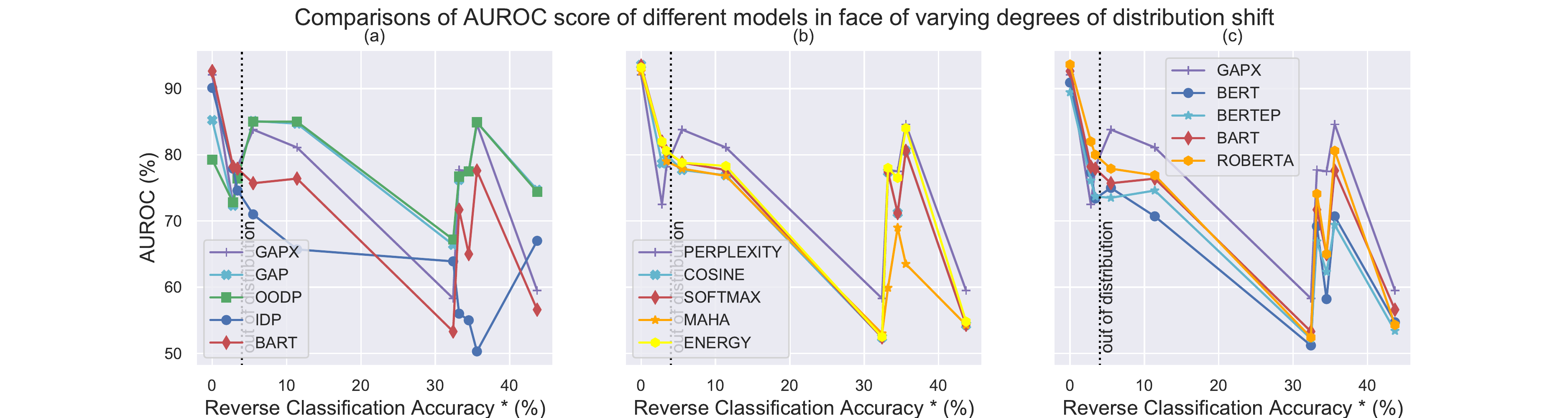}
    \caption{Comparing AUROC scores of different models at varying degrees of distribution shift.}
    \label{aucplots}
\end{figure}

\section{Additional Ablations}
\subsection{Additional Variants}
We also conducted ablations on several variants of GAPX.
\begin{itemize}
    \item \textbf{GAPX(neg-log)} replaces perplexity with the neg-log likelihood of the concatenated $(s_1, s_2)$ as the measure. Specifically, GAPX(neg-log) modifies the Eqn.~\ref{perplexity} and Eqn.~\ref{lambda definition} such that:
    \begin{align*}
        NLL(s_1, s_2) &= \log ((\prod_{i=1}^{n}\frac{1}{P(w^i|w^{1:i-1})}).\\
        \lambda(s_1, s_2) &=  cdf(NLL(s_1,s_2), Weibull(a, c, loc, scale)).
    \end{align*}
    
    \item \textbf{GAPX(w/ IDP)} ensembles GAP with IDP.
    
    \item \textbf{GAPX(w/ BART)} explores the option of ensembling GAP with BART as described in Appendix.\ref{sec:baselines}.
    
\end{itemize}
These variants of GAPX mostly perform similarly to the GAPX model proposed in the main paper, which ensembles GAP with RoBERTa.
\begin{table*}[!h]
\centering
\resizebox{\columnwidth}{!}{%
\begin{tabular}{lcccccccc}
Model  & QQP & PIT & PIT & PAWS & PAWS & PAWS & QQP& \multirow{2}{*}{average}\\
 & -> WMT   & -> WMT & -> PAWS & -> QQP & -> PIT & -> WMT  &-> PAWS \\
 & (5.5) & (11.4) & (32.4) & (33.2) & (34.5) & (35.6)  & (43.7)\\
 \hline
 GAPX & 75.5/75.5 & 74.4/74.5 & 55.1/55.5 & 70.8/70.8 & 69.7/70.2 & 76.4/76.4 & 52.3/54.3 & 67.7/68.2\\
GAPX(neg-log) & 75.9/75.9 & 74.8/74.8 & 56.0/61.7 & 62.6/65.9 & 62.9/63.4 & 72.3/72.4 & 62.9/63.0 & 66.8/68.2\\
GAP(w/ IDP)   & 73.2/73.4 & 74.0/74.0  & 55.4/61.6 & 57.8/58.3 & 54.7/54.7 & 67.7/67.9& 66.3/66.5 & 64.2/65.2\\
GAPX(w/ BART) & 74.6/74.6 &74.6/74.6  & 55.2/55.4 & 70.7/70.8 &69.7/70.2 & 75.9/76.0 & 55.1/56.0 & 68.0/68.2\\
GAPX(w/ BERT) & 72.6/72.6 & 74.7/74.7 & 55.1/55.2 &70.2/70.2  &69.7/70.2 & 75.7/75.9 & 56.1/56.3 & 67.7/67.9\\
\end{tabular}
}
\caption{Performance on different out-of-distribution combinations of QQP, PAWS and PIT, in terms of macro F1/accuracy (ACC). }
\label{table5}
\end{table*}

\begin{table}[!h]
\centering
\begin{tabular}{lccccccr}
Model & QQP & PAWS  & PIT & QQP & PIT& \multirow{2}{*}{average}\\
& ->QQP & ->PAWS  & ->PIT & ->PIT & ->QQP&\\
& (0) & (0)  & (0) & (2.8) & (3.4) &\\
\hline
GAPX & 84.4/84.5 & 92.7/92.7& 79.3/79.3& 62.3/63.4 & 72.0/72.4 & 78.1/78.5\\
GAPX(neg-log) & 83.3/83.4 & 91.2/91.5& 80.9/80.9& 68.9/70.1 & 72.2/72.7 & 79.3/78.7\\
GAP(threshed) & 78.2/78.3 & 86.5/86.8 & 77.4/77.7 & 67.4/68.5 & 69.4/69.8 & 75.8/76.2\\
GAPX(w/ BART) & 82.6/82.8 & 93.0/93.1 & 79.4/79.4 & 62.3/63.4& 71.4/71.5& 77.7/78.0\\
GAPX(w/ BERT) & 82.6/82.7 &91.9/92.0 & 76.8/76.8 & 62.3/63.4 &69.9/69.9 & 76.7/77.0\\
\end{tabular}
\caption{In-distribution performances on QQP, PAWS, and PIT, in terms of macro F1/accuracy (ACC).}
\label{table6}
\end{table}
\begin{table*}[!h]
\centering
\resizebox{\columnwidth}{!}{%
\begin{tabular}{lcccccccc}
Model  & QQP & PIT & PIT & PAWS & PAWS & PAWS & QQP & \multirow{2}{*}{average}\\
 & -> WMT   & -> WMT & -> PAWS & -> QQP & -> PIT & -> WMT  &-> PAWS & \\
 & (5.5) & (11.4) & (32.4) & (33.2) & (34.5) & (35.6)  & (43.7)\\
 \hline
GAPX(perplexity)  & 75.5/75.5 & 74.4/74.5 & 55.1/55.5 & 70.8/70.8 & 69.7/70.2 & 76.4/76.4 & 52.3/54.3 & 67.7/68.2\\
GAPX(cosine) & 65.2/66.9 & 53.7/59.4 &31.2/45.4  & 60.0/63.5 & 63.2/63.9 & 59.8/62.8 & 35.4/46.7 & 52.7/58.4\\
GAPX(softmax) &70.3/71.1 & 63.2/65.7 &31.2/45.4 & 60.0/63.5 &  63.2/63.9&59.8/62.8  & 39.2/48.2 & 55.3/60.1\\
GAPX(maha) &65.3/66.9 &60.5/63.6  &35.7/46.8 & 63.2/64.8 & 65.5/65.7 & 61.0/63.0 & 35.1/46.5 & 55.2/59.6\\
GAPX(energy)  &69.9/70.8 & 66.6/68.2 &31.4/45.4  & 70.7/70.8 & 69.5/70.0 & 76.7/76.7 & 39.5/48.4& 60.6/64.3\\
\end{tabular}
}
\caption{Out-of-distribution performance when using different out-of-distribution metrics.}
\label{table7}
\end{table*}

\begin{table}[!h]
\centering
\begin{tabular}{lcccccr}
Model & QQP & PAWS  & PIT & QQP & PIT& \multirow{2}{*}{average}\\
& ->QQP & ->PAWS  & ->PIT & ->PIT & ->QQP\\
& (0) & (0)  & (0) & (2.8) & (3.4) \\
\hline
GAPX(perplexity) & 84.4/84.5 & 92.7/92.7& 79.3/79.3& 62.3/63.4 & 72.0/72.4 & 78.1/78.5  \\
GAPX(cosine) &82.6/82.8 & 93.5/93.5 &81.0/81.1 & 66.0/67.2 & 73.0/74.0 &  79.2/79.7\\
GAPX(softmax) &82.2/82.5 & 93.5/93.5 &82.6/82.6 & 71.9/72.4 &  73.8/74.6 & 80.8/81.1\\
GAPX(maha) &84.4/84.5 &92.2/92.2  &81.3/81.3 & 66.4/68.6 &  72.5/73.2 & 79.4/80.0\\
GAPX(energy) &82.8/83.0 &92.2/92.2  &81.4/81.5 & 71.7/72.3 &  73.6/74.3& 80.3/80.7\\
\end{tabular}
\caption{In-distribution performance of different out-of-distribution metrics.}
\label{table8}
\end{table}
\section{Additional Discussions}
\subsection{Ablations on $C$, Eqn.~\ref{ood emsemble}}
\begin{table*}[!h]
\centering
\resizebox{\columnwidth}{!}{%
\begin{tabular}{lccccccc}
Model  & QQP & PIT & PIT & PAWS & PAWS & QQP& \multirow{2}{*}{average}\\
 &-> PIT & -> QQP & -> PAWS & -> QQP & -> PIT &-> PAWS \\
 &  (2.8) &  (3.4) & (32.4) & (33.2) & (34.5)   & (43.7)\\
 \hline
GAPX(0) & 62.3/63.4 & 72.0/72.4 & 55.1/55.5 & 70.8/70.8 & 69.7/70.2 &52.3/54.3 & 63.7/64.4\\
GAPX(10) & 63.3/64.0 & 72.1/72.7 & 48.3/51.5 & 69.5/69.5 & 69.7/70.2 & 47.3/52.4 & 61.7/63.4\\
GAPX(100) & 66.9/67.1 & 72.1/72.6 & 48.4/50.7 &71.3/71.5  & 71.2/71.3 & 52.3/54.4 & 63.7/64.6\\
\end{tabular}
}
\caption{Comparing using different amount of data to set $C$. Macro F1/ACC are reported.}
\label{table9}
\end{table*}

To set $C$ in Equation.~\ref{ood emsemble}, in the case of 0 samples, we roughly estimate an integer from 1-5 for the interpolation constant $C$. For adversarial distributions like PAWS, we expect a lower perplexity because both sentences share the same bag-of-word, so we set $C = 1$. For standard distributions like QQP and WMT, we expect a modest perplexity, so we set $C = 3$. 
For informal distributions like PIT, where sentences do not strictly follow syntax and grammar, we expect a higher perplexity, so we set $C = 5$. 
In other cases, with validation data, we determine the best constant $C$ based on the validation data (if there are multiple constants with the same results on the validation data, we take the smallest one). As shown in Table~\ref{table3} (WMT results are not included here due to the lack of validation data), although using 0 or 10 samples achieve worse performances than using 100 samples, meaning the best constant threshold is not found, the results appear to be stable overall (on average 2\% fluctuations in macro F1 and 1\% fluctuations in accuracy). If a small amount of validation data is accessible, the performance of GAPX can be further improved.

\subsection{Interpreting the Results}
\begin{figure}[h!]
    \centering
    \includegraphics[width=.5\textwidth]{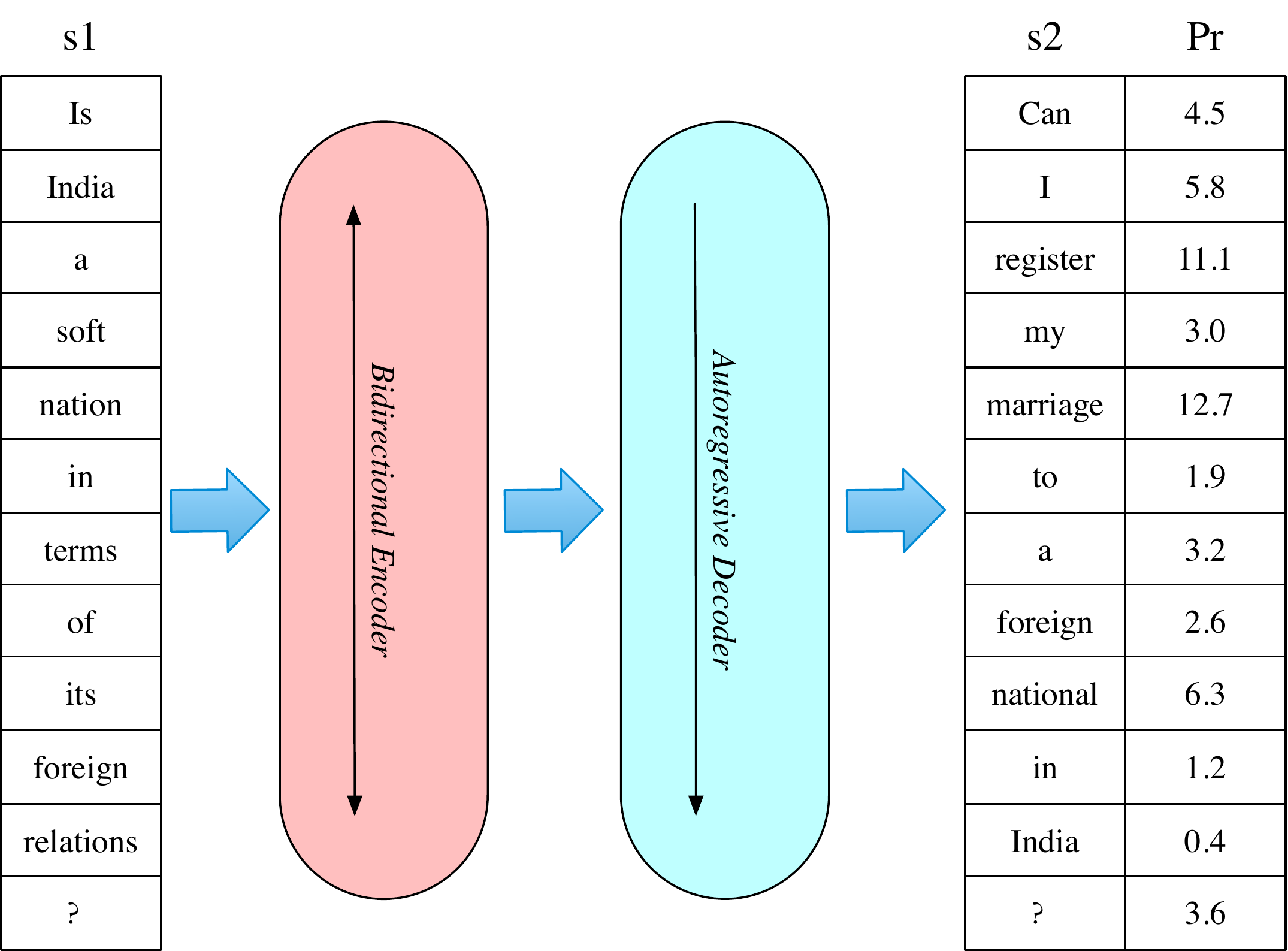}
    \caption{An example from QQP illustrating how to interpret the result of our method, by OODP.}
    \label{OODP}
\end{figure}
\begin{figure}[h!]
    \centering
    \includegraphics[width=.5\textwidth]{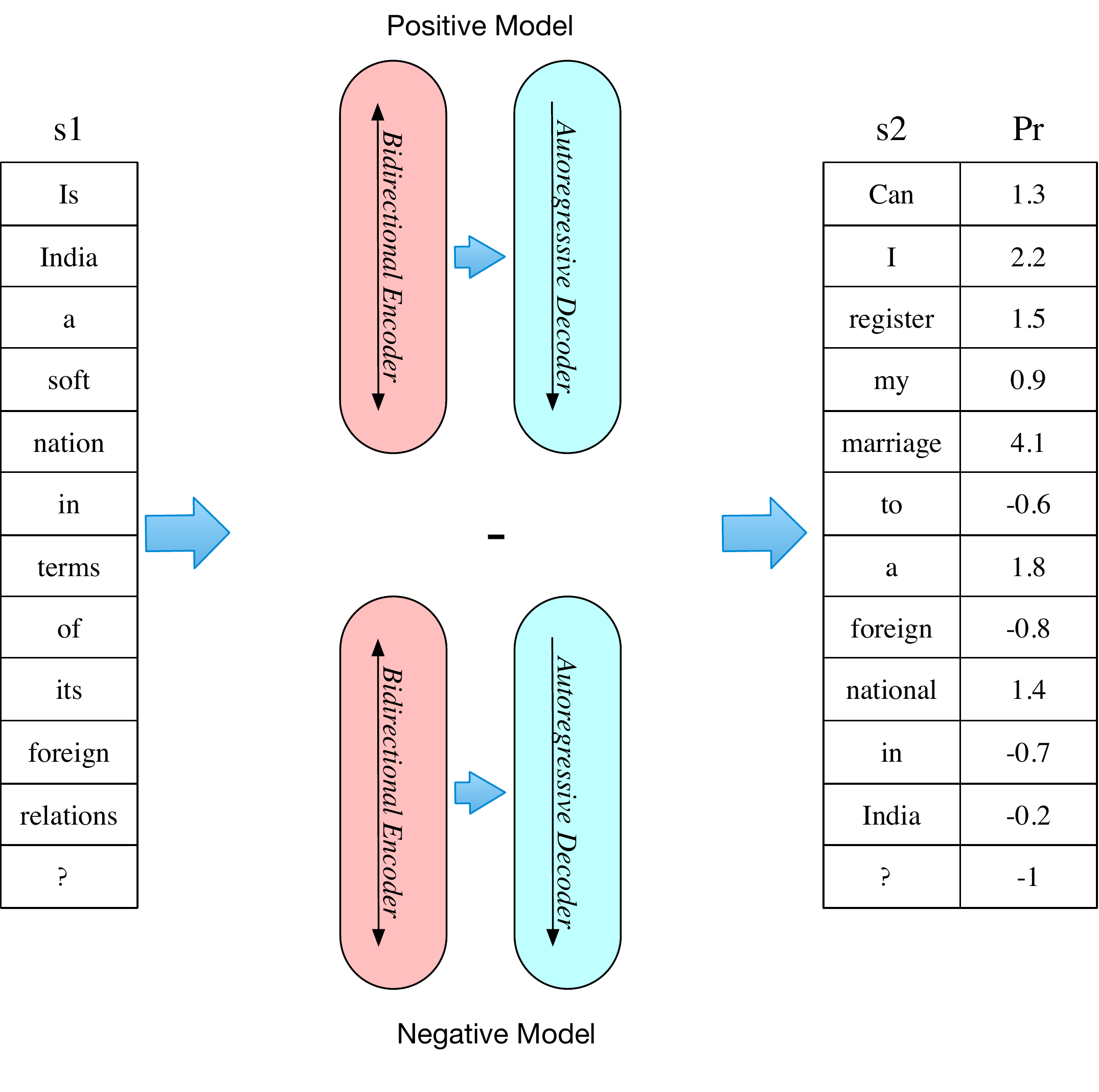}
    \caption{An example from QQP illustrating how to interpret the result of our method, by IDP.}
    \label{IDP}
\end{figure}
\begin{figure}[h!]
    \centering
    \includegraphics[width=.5\textwidth]{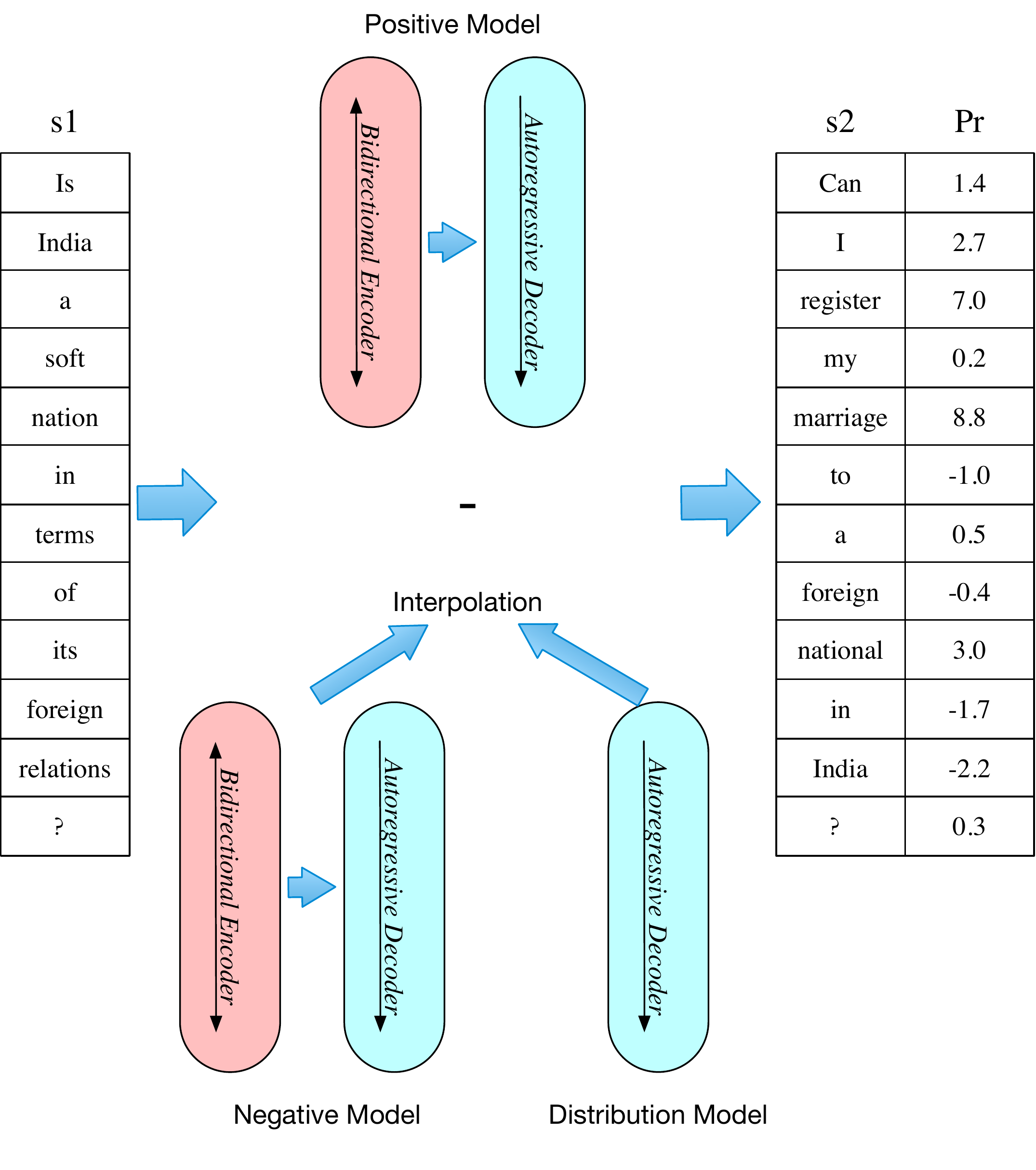}
    \caption{An example from QQP illustrating how to interpret the result of our method, by GAP.}
    \label{GAP}
\end{figure}
Figure~\ref{OODP}, Figure~\ref{IDP}, and Figure~\ref{GAP} shows examples of how our autoregressive paraphrase identification models work. For OODP, our model will output a log of conditional probability for each word in $s_2$ given $s_1$ and all the previous words in $s_2$, namely:

{\small
\begin{align*}
    \log P(w_2^{(i)}|s_1, Y = 1, w_2^{(1:i-1)}).
\end{align*}
}

For IDP, we can use the log of the quotient of the conditional probability for each word given by the positive model and the negative model as an indicator which words contribute the most to the prediction result:

{\small
\begin{align*}
    \log P(w_2^{(i)}|s_1, Y = 1, w_2^{(1:i-1)}) - \log P(w_2^{(i)}|s_1, Y = 0, w_2^{(1:i-1)}).
\end{align*}
}

For GAP, we can use the the score defined in Eqn..~\ref{ood emsemble} split on each word, namely:

{\small
\begin{align*}
    \log P(w_2^{(i)}|s_1, Y = 1, w_2^{(1:i-1)}) - (1 - \lambda(s_1, s_2))\log P(w_2^{(i)}|s_1, Y = 0, w_2^{(1:i-1)}) - \lambda(s_1, s_2) C.
\end{align*}
}

In all three models, higher scores represent a higher chance of being non-paraphrases. For IDP and GAP, the threshold is 0 while for OODP the threshold is 3. All three models predict this sentence pair to be non-paraphrases, attending to slightly different key words. The top 3 words with the highest scores in OODP and GAP are 'register', 'marriage', and 'national'. All of them represent the words that are unlikely to occur in a paraphrase of the original sentence. The top 3 words with the highest scores in IDP are 'I', 'marriage', and 'a'. Its reliance on the word 'a' might be due to the error propagation of the autoregressive decoding.

\subsection{Implementation of out-of-distribution metrics}
To compare our perplexity metric with different off-the-shelf out-of-distribution metrics (Fig.~\ref{accplots}(b) and \ref{aucplots}(b)), we first train a RoBERTa model as described in Sec.~\ref{sec:baselines}. Both MAHA and COSINE need in-distribution validation data, for which we use half of the development data provided in each dataset. We use the other half of the development data to estimate the Weibull distribution of the metrics. We rely on the implementation of SOFTMAX, ENERGY, MAHA, and COSINE from \citet{DBLP:journals/corr/abs-2104-08812}. The metrics are calculated as follow:
\begin{enumerate}
    \item \textbf{SOFTMAX}. We use the maximum class probability $1 - max_{j=0,1} pj$ among 2 classes (paraphrases and non-paraphrases) in the final softmax layer.
    \item \textbf{ENERGY}. We use the following formula to calculate energy score:
    \begin{align*}
        g = -\log \sum_{j = 0}^1 \exp{(w_j^Th)},
    \end{align*}
    where $w_j$ is the weight of the $j$th class in the softmax layer, and $h$ is the input to the softmax layer (of the concatenated $(s_1, s_2)$ input).
    
    \item \textbf{MAHA}. We use the input representation $h$ of the penultimate layer of the model, and fit a Gaussian distribution to each class in the in-distribution development data $\mathcal{D}_{val} = \{(x_i, y_i)\}_{i = 1}^M$:
    \begin{align*}
        \mu_j &= \mathbf{E}_{y_i = j}[h_j]\\
        \Sigma &= \mathbf{E}[(h_i - \mu_{y_i})(h_i - \mu_{y_i})^T].\\
    \end{align*}
    Then, the MAHA distance is calculated as:
    \begin{align*}
        g = -\min_{j = 0, 1} (h - \mu_j)\Sigma^+(h - \mu_j),
    \end{align*}
    where $\Sigma^+$ is the pseudo-inverse of $\Sigma$.
    
    \item \textbf{COSINE}. We use the maximum cosine similarity of $h$ (of the concatenated $(s_1, s_2)$ input) to samples in the validation dataset:
    \begin{align*}
        g = -\max_{i = 1}^{M} \cos{(h, h_i^{(val)})}.
    \end{align*}
    
\end{enumerate}

\subsection{RCA Implementation Details}
To instantiate a measurement of RCA* scores. We use the RoBERTa model described in Section~\ref{sec:baselines} as our classifier. To measure the distribution shift from a source distribution $\mathcal{D}^s$ to a target distribution $\mathcal{D}^t$. We use a training set of $\mathcal{D}^s$, a test set of $\mathcal{D}^s$, and a development set of $\mathcal{D}^t$. All the measurements of RCA* scores in this paper fix the size of the development to be 1000, and use the test set of $\mathcal{D}^s$ to be the entire test set of the original dataset. 

\begin{figure}[h!]
    \centering
    \includegraphics[width=\textwidth]{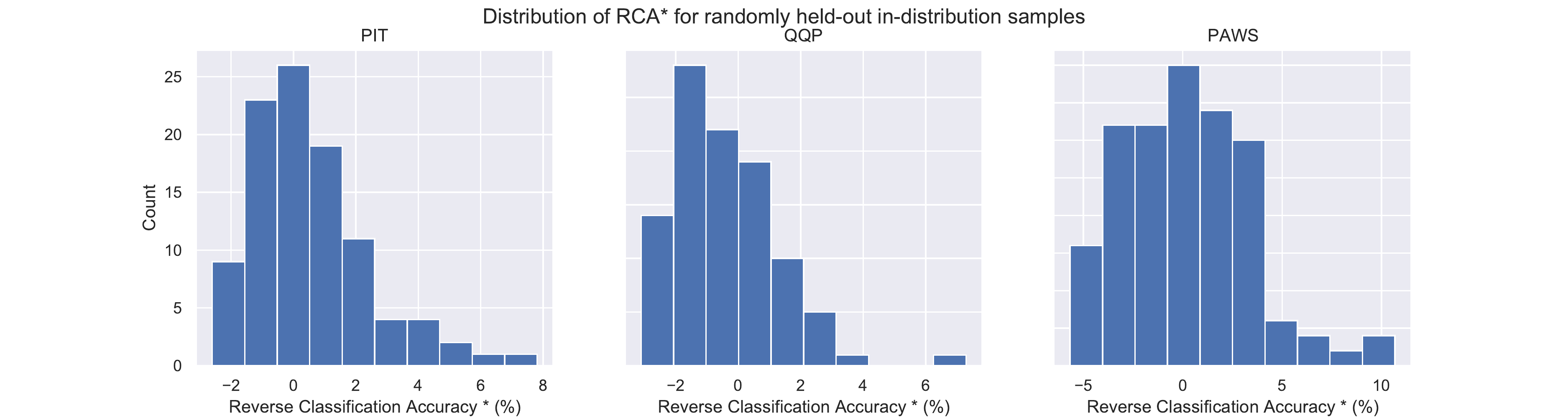}
    \caption{Distribution of RCA* for randomly held-out in-distribution samples}
    \label{rca_distribution}
\end{figure}

To calibrate the RCA* scores, we measure the distribution shift of one dataset to itself. We first train a binary classifier RoBERTa $M1$ (of paraphrases and non-paraphrase) with the training set of $\mathcal{D}^s$ for 3 epoches with an Adam Optimizer of learning rate 2e-5 (training with multiple runs with different random seeds to select the best model on a development set of $\mathcal{D}^s$, different from the development set of $\mathcal{D}^t$). Then we apply $M1$ to the development set of $\mathcal{D}^t$ to relabel those data. We take the relabeled data to retrain a classifier $M2$ and apply $M2$ to the test set of $\mathcal{D}^s$. We measure the performance drop from $M1$ to $M2$ on the test set of $\mathcal{D}^s$ as the RCA score. Note here that we make small changes to the originally proposed RCA score where the performances are measured in terms of ACC scores. We found that AUROC scores are in practice more stable for measuring RCA scores, so we use AUROC scores instead. To get the final RCA* score, we use the equation:
\begin{align*}
    RCA*(\mathcal{D}^s, \mathcal{D}^t) = RCA(\mathcal{D}^s, \mathcal{D}^t) - RCA(\mathcal{D}^s, \mathcal{D}^s),
\end{align*}
where $RCA*(\mathcal{D}^s, \mathcal{D}^t)$ is the RCA* score from $\mathcal{D}^s$ to $\mathcal{D}^t$, while $RCA(\mathcal{D}^s, \mathcal{D}^t)$ is the RCA score from $\mathcal{D}^s$ to $\mathcal{D}^t$. Under this definition, the RCA* score of one dataset to itself is defined to be 0. To repeat the measurements of distribution shift from $\mathcal{D}^s$ to itself, at each time, we hold-back 1000 random pairs from $\mathcal{D}^s$, and measure the RCA. We include plots of the distribution of the RCA* in Fig.~\ref{rca_distribution} as well as the corresponding raw RCA* data below:

PIT: 4.4, 0.65, 1.9, 0.91, 5.1, 4.0, 1.1, 0.63, -0.1, -0.43, 2.9, 0.6, -1.7, 0.83, -0.27, 7.8, 1.7, 3.8, 0.78, -1.2, 2.1, 1.1, -1.3, -1.2, 1.3, -1.2, 0.032, 3.0, -2.0, -1.8, -1.2, 1.8, -2.4, 6.2, -0.47, 0.26, -0.88, -1.1, 4.2, 3.4, 3.3, 0.22, -0.25, 0.065, 2.6, 0.15, -0.93, 0.27, -0.49, -1.5, -0.79, 0.38, 2.3, 0.83, -1.4, 0.81, -0.63, -1.6, 0.97, -1.6, 0.82, 1.1, 1.9, 2.3, 0.23, -1.1, 0.72, -0.87, 1.7, 0.04, -0.38, 0.23, 1.7, -1.2, 0.1, 1.4, -0.15, 1.2, -2.6, -1.4, 0.75, 2.3, -1.2, -0.68, 5.6, -0.13, -1.1, 0.16, -0.39, 0.097, -1.9, -1.1, -0.26, -1.3, -1.3, -0.8, 0.3, 0.73, -1.9, -0.51

QQP: -0.27, 0.72, -2.3, 1.3, -2.5, 1.1, -1.1, -1.4, 2.0, -2.1, -0.77, -0.63, 0.045, -0.97, -2.1, 0.11, 1.3, -0.25, -1.1, 0.39, 0.65, 1.2, 0.87, 1.5, 0.058, 3.5, -0.039, 1.6, -1.2, 2.5, 0.64, -0.48, 2.5, -1.8, -0.52, -2.1, -0.22, -1.4, 0.97, 3.0, -2.9, -0.92, -0.42, 0.72, 1.8, -1.3, 0.63, -2.0, 0.4, -1.2, -0.65, -2.0, 0.7, 0.95, -1.3, -2.1, 1.2, -1.0, 2.5, -0.32, -1.8, -0.59, -0.016, -1.4, -1.3, 2.5, 1.4, -2.2, -1.2, 0.98, 0.93, 0.98, -2.1, -1.1, -1.8, -0.43, -0.42, 7.3, -0.75, -1.5, -0.87, -2.4, -0.61, 0.084, -1.7, -0.16, -2.4, -2.0, -1.6, -2.9, -1.7, -2.3, -1.9, -2.3, -1.7, -3.1, -1.6, 0.2, -1.7, -2.9

PAWS: -3.0, -0.0092, -0.46, -5.6, -1.7, 2.7, 0.5, 2.4, 3.6, 0.034, -4.1, -3.2, 3.6, 8.2, 0.71, -0.079, 2.9, 5.4, 1.7, -5.2, 3.2, 2.8, -3.0, 2.0, -2.2, -1.7, -0.2, 4.1, -0.22, 3.9, 1.1e+01, 1e+01, 1.2, 0.0074, -0.91, -2.2, 3.9, -2.9, -2.7, -2.8, -4.2, -4.1, -1.0, -3.1, 2.6, -0.83, 0.74, 1.7, 0.56, 2.2, 0.87, -0.5, 0.83, -5.5, 0.4, -3.0, 1.1, -3.1, 0.92, -1.6, 4.1, -3.1, -0.99, 0.79, -0.15, -1.3, 5.2, 0.18, -5.7, 3.4, 2.9, 3.2, -4.0, -2.4, -4.9, -0.51, -1.8, -2.7, -3.7, 6.6, 0.65, 2.5, 1.5, -2.3, 3.1, 1.9, 1.2, -2.1, 1.7, 0.95, 2.1, -2.7, 5.9, -4.0, 1.8, -1.8, -0.2, 4.6, -1.3, -1.0

\end{document}